%% file: dilation.tex
\documentclass{article} 
\usepackage[toc,title]{appendix}
\usepackage{iclr2016_conference,times}
\usepackage{hyperref}
\usepackage{url}
\usepackage{epsfig}
\usepackage{graphicx}
\usepackage{amsmath}
\usepackage{amssymb}
\usepackage{subcaption}
\usepackage{verbatim}
\usepackage{caption}
\usepackage{natbib}
\usepackage{adjustbox}
\usepackage{array}

\iclrfinalcopy

\def\pp{\mathbf p}

\def\ss{\mathbf s}
\def\tt{\mathbf t}

\def\N{\mathcal N}

\def\Re{\mathbb{R}}
\def\Ze{\mathbb{Z}}

\newcommand\ver[1]{\rotatebox[origin=c]{90}{#1}}

\newcommand{\timess}{\mathbin{\!\times\!}}

\title{Multi-Scale Context Aggregation by\\Dilated Convolutions}

\author{Fisher Yu\\
Princeton University
\AND
Vladlen Koltun\\
Intel Labs
}

\begin{document}

\maketitle

\begin{abstract}
State-of-the-art models for semantic segmentation are based on adaptations of convolutional networks that had originally been designed for image classification. However, dense prediction problems such as semantic segmentation are structurally different from image classification. In this work, we develop a new convolutional network module that is specifically designed for dense prediction. The presented module uses dilated convolutions to systematically aggregate multi-scale contextual information without losing resolution. The architecture is based on the fact that dilated convolutions support exponential expansion of the receptive field without loss of resolution or coverage. We show that the presented context module increases the accuracy of state-of-the-art semantic segmentation systems. In addition, we examine the adaptation of image classification networks to dense prediction and show that simplifying the adapted network can increase accuracy.
\end{abstract}

\section{Introduction}

Many natural problems in computer vision are instances of dense prediction. The goal is to compute a discrete or continuous label for each pixel in the image. A prominent example is semantic segmentation, which calls for classifying each pixel into one of a given set of categories \citep{He2004,Shotton2009,Kohli2009,KrahenbuhlKoltun2011}. Semantic segmentation is challenging because it requires combining pixel-level accuracy with multi-scale contextual reasoning \citep{He2004,GalleguillosBelongie2010}.

Significant accuracy gains in semantic segmentation have recently been obtained through the use of convolutional networks \citep{LeCun1989} trained by backpropagation \citep{Rumelhart1986}.
Specifically, \cite{Long2015} showed that convolutional network architectures that had originally been developed for image classification can be successfully repurposed for dense prediction. These reporposed networks substantially outperform the prior state of the art on challenging semantic segmentation benchmarks.
This prompts new questions motivated by the structural differences between image classification and dense prediction.
Which aspects of the repurposed networks are truly necessary and which reduce accuracy when operated densely?
Can dedicated modules designed specifically for dense prediction improve accuracy further?

Modern image classification networks integrate multi-scale contextual information via successive pooling and subsampling layers that reduce resolution until a global prediction is obtained \citep{Krizhevsky2012,SimonyanZisserman2015}.
In contrast, dense prediction calls for multi-scale contextual reasoning in combination with full-resolution output. Recent work has studied two approaches to dealing with the conflicting demands of multi-scale reasoning and full-resolution dense prediction. One approach involves repeated up-convolutions that aim to recover lost resolution while carrying over the global perspective from downsampled layers \citep{Noh2015,Fischer2015}. This leaves open the question of whether severe intermediate downsampling was truly necessary. Another approach involves providing multiple rescaled versions of the image as input to the network and combining the predictions obtained for these multiple inputs \citep{Farabet2013,Lin2015,Chen2015Scale}. Again, it is not clear whether separate analysis of rescaled input images is truly necessary.

In this work, we develop a convolutional network module that aggregates multi-scale contextual information without losing resolution or analyzing rescaled images. The module can be plugged into existing architectures at any resolution. Unlike pyramid-shaped architectures carried over from image classification, the presented context module is designed specifically for dense prediction.
It is a rectangular prism of convolutional layers, with no pooling or subsampling. The module is based on dilated convolutions, which support exponential expansion of the receptive field without loss of resolution or coverage.

As part of this work, we also re-examine the performance of repurposed image classification networks on semantic segmentation. The performance of the core prediction modules can be unintentionally obscured by increasingly elaborate systems that involve structured prediction, multi-column architectures, multiple training datasets, and other augmentations. We therefore examine the leading adaptations of deep image classification networks in a controlled setting and remove vestigial components that hinder dense prediction performance. The result is an initial prediction module that is both simpler and more accurate than prior adaptations.

Using the simplified prediction module, we evaluate the presented context network through controlled experiments on the Pascal VOC 2012 dataset \citep{Everingham2010}. The experiments demonstrate that plugging the context module into existing semantic segmentation architectures reliably increases their accuracy.

\section{Dilated Convolutions}
\label{sec:convolutions}

Let $F: \Ze^2 \rightarrow \Re$ be a discrete function. Let $\Omega_r = [-r,r]^2 \cap \Ze^2$ and let \mbox{$k: \Omega_r \rightarrow \Re$} be a discrete filter of size $(2r+1)^2$. The discrete convolution operator $\ast$ can be defined as
\begin{equation}
(F \ast k)(\pp) = \sum_{\ss + \tt = \pp} F(\ss) \, k(\tt).
\label{eq:regular}
\end{equation}
We now generalize this operator. Let $l$ be a dilation factor and let $\ast_l$ be defined as
\begin{equation}
(F \ast_l k)(\pp) = \sum_{\ss + l \tt = \pp} F(\ss) \, k(\tt).
\label{eq:dilated}
\end{equation}
We will refer to $\ast_l$ as a dilated convolution or an $l$-dilated convolution. The familiar discrete convolution $\ast$ is simply the $1$-dilated convolution.

The dilated convolution operator has been referred to in the past as ``convolution with a dilated filter". It plays a key role in the {\em algorithme \`{a} trous}, an algorithm for wavelet decomposition \citep{Holschneider1987,Shensa1992}.\footnote{Some recent work mistakenly referred to the dilated convolution operator itself as the {\em algorithme \`{a} trous}. This is incorrect. The {\em algorithme \`{a} trous} applies a filter at multiple scales to produce a signal decomposition. The algorithm uses dilated convolutions, but is not equivalent to the dilated convolution operator itself.} We use the term ``dilated convolution" instead of ``convolution with a dilated filter" to clarify that no ``dilated filter" is constructed or represented. The convolution operator itself is modified to use the filter parameters in a different way. The dilated convolution operator can apply the same filter at different ranges using different dilation factors. Our definition reflects the proper implementation of the dilated convolution operator, which does not involve construction of dilated filters.


In recent work on convolutional networks for semantic segmentation, \cite{Long2015} analyzed filter dilation but chose not to use it. \cite{Chen2015ICLR} used dilation to simplify the architecture of \cite{Long2015}.
In contrast, we develop a new convolutional network architecture that systematically uses dilated convolutions for multi-scale context aggregation.

Our architecture is motivated by the fact that dilated convolutions support exponentially expanding receptive fields without losing resolution or coverage. Let $F_0, F_1, \ldots , F_{n-1}: \Ze^2 \rightarrow \Re$ be discrete functions and let \mbox{$k_0, k_1, \ldots , k_{n-2}: \Omega_1 \rightarrow \Re$} be discrete $3\timess 3$ filters. Consider applying the filters with exponentially increasing dilation:
\begin{eqnarray}
F_{i+1} = F_{i} \ast_{2^{i}} k_{i} \quad \textup{for\ \ } i = 0, 1, \ldots, n-2.
\label{eq:exponential}
\end{eqnarray}

\begin{figure}[t]
  \begin{center}
    \begin{subfigure}[b]{0.3\linewidth}
        \includegraphics[width=\textwidth]{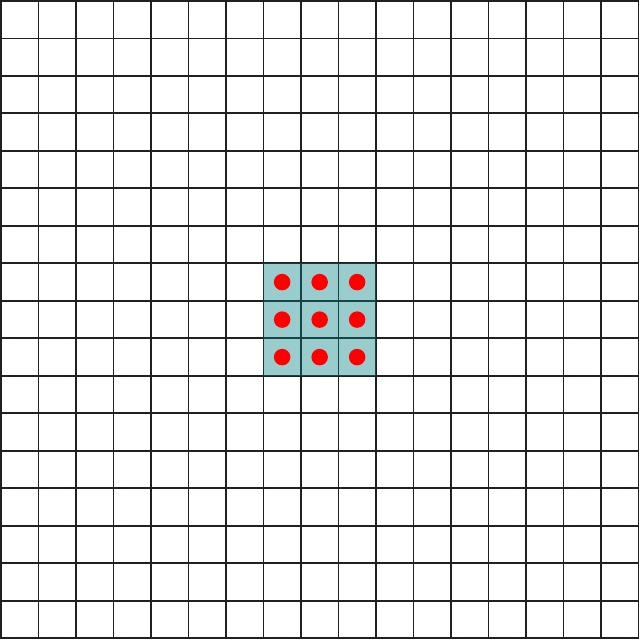}
        \caption{}
    \end{subfigure}
    ~~
    \begin{subfigure}[b]{0.3\linewidth}
        \includegraphics[width=\textwidth]{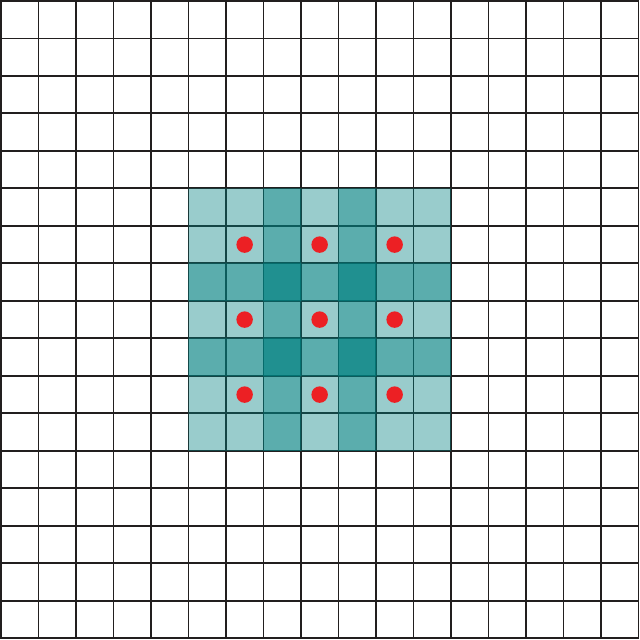}
        \caption{}
    \end{subfigure}
    ~~
    \begin{subfigure}[b]{0.3\linewidth}
        \includegraphics[width=\textwidth]{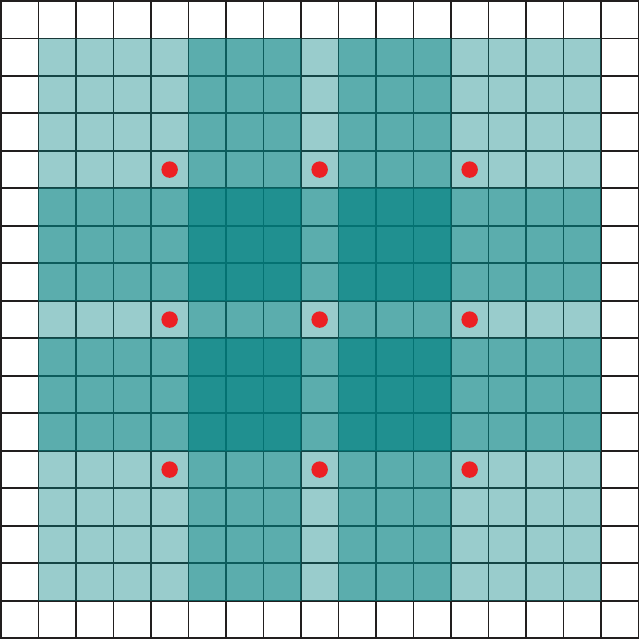}
        \caption{}
    \end{subfigure}
  \end{center}
  \vspace{-2mm}
  \caption{Systematic dilation supports exponential expansion of the receptive field without loss of resolution or coverage. (a) $F_1$ is produced from $F_0$ by a 1-dilated convolution; each element in $F_1$ has a receptive field of $3\timess 3$. (b) $F_2$ is produced from $F_1$ by a 2-dilated convolution; each element in $F_2$ has a receptive field of $7\timess 7$. (c) $F_3$ is produced from $F_2$ by a 4-dilated convolution; each element in $F_3$ has a receptive field of $15\timess 15$. The number of parameters associated with each layer is identical. The receptive field grows exponentially while the number of parameters grows linearly.}
  \label{fig:exponential}
  \vspace{-2mm}
\end{figure}

Define the receptive field of an element $\pp$ in $F_{i+1}$ as the set of elements in $F_0$ that modify the value of $F_{i+1}(\pp)$. Let the size of the receptive field of $\pp$ in $F_{i+1}$ be the number of these elements. It is easy to see that the size of the receptive field of each element in $F_{i+1}$ is \mbox{$(2^{i+2} - 1)\timess(2^{i+2} - 1)$}. The receptive field is a square of exponentially increasing size. This is illustrated in Figure~\ref{fig:exponential}.

\section{Multi-Scale Context Aggregation}
\label{sec:context}

The context module is designed to increase the performance of dense prediction architectures by aggregating multi-scale contextual information.
The module takes $C$ feature maps as input and produces $C$ feature maps as output. The input and output have the same form, thus the module can be plugged into existing dense prediction architectures.

We begin by describing a basic form of the context module. In this basic form, each layer has $C$ channels. The representation in each layer is the same and could be used to directly obtain a dense per-class prediction, although the feature maps are not normalized and no loss is defined inside the module. Intuitively, the module can increase the accuracy of the feature maps by passing them through multiple layers that expose contextual information.

The basic context module has 7 layers that apply $3\timess 3$ convolutions with different dilation factors. The dilations are 1, 1, 2, 4, 8, 16, and 1. Each convolution operates on all layers: strictly speaking, these are $3\timess 3\timess C$ convolutions with dilation in the first two dimensions. Each of these convolutions is followed by a pointwise truncation $\max(\cdot,0)$. A final layer performs $1\timess 1\timess C$ convolutions and produces the output of the module. The architecture is summarized in Table \ref{tab:layers}. Note that the front-end module that provides the input to the context network in our experiments produces feature maps at $64\timess 64$ resolution. We therefore stop the exponential expansion of the receptive field after layer 6.

Our initial attempts to train the context module failed to yield an improvement in prediction accuracy. Experiments revealed that standard initialization procedures do not readily support the training of the module. Convolutional networks are commonly initialized using samples from random distributions \citep{GlorotBengio2010,Krizhevsky2012,SimonyanZisserman2015}. However, we found that random initialization schemes were not effective for the context module.
We found an alternative initialization with clear semantics to be much more effective:
\begin{equation}
k^b (\tt, a) = 1_{[\tt = 0]}1_{[a=b]},
\end{equation}
where $a$ is the index of the input feature map and $b$ is the index of the output map. This is a form of identity initialization, which has recently been advocated for recurrent networks \citep{Le2015}. This initialization sets all filters such that each layer simply passes the input directly to the next. A natural concern is that this initialization could put the network in a mode where backpropagation cannot significantly improve the default behavior of simply passing information through. However, experiments indicate that this is not the case. Backpropagation reliably harvests the contextual information provided by the network to increase the accuracy of the processed maps.

\begin{table}
\centering
\small
\begin{tabular}{|l|c|c|c|c|c|c|c|c|}
\hline
Layer 							& 1 & 2 & 3 & 4 & 5 & 6 & 7 & 8 \\ \hline \hline
Convolution 					& $3\timess 3$ & $3\timess 3$ & $3\timess 3$ & $3\timess 3$ & $3\timess 3$ & $3\timess 3$ & $3\timess 3$ & $1\timess 1$ \\ \hline
Dilation	                  	& 1 & 1 & 2 & 4 & 8 & 16 & 1 & 1 \\ \hline
Truncation					  	& Yes & Yes & Yes & Yes & Yes & Yes & Yes & No \\ \hline
Receptive field	               	& $3\timess 3$ & $5\timess 5$ & $9\timess 9$ & $17\timess 17$ & $33\timess 33$ & $65\timess 65$ & $67\timess 67$ & $67\timess 67$ \\ \hline \hline
\multicolumn{9}{|l|}{Output channels}				    			\\ \hline
Basic	                  	& $C$ & $C$ & $C$ & $C$ & $C$ & $C$ & $C$ & $C$ \\ \hline
Large	                  	& $2C$ & $2C$ & $4C$ & $8C$ & $16C$ & $32C$ & $32C$ & $C$ \\ \hline
\end{tabular}
\caption{Context network architecture. The network processes $C$ feature maps by aggregating contextual information at progressively increasing scales without losing resolution.}
\label{tab:layers}
\vspace{-3mm}
\end{table}

This completes the presentation of the basic context network. Our experiments show that even this basic module can increase dense prediction accuracy both quantitatively and qualitatively. This is particularly notable given the small number of parameters in the network: $\approx\! 64C^2$ parameters in total.

We have also trained a larger context network that uses a larger number of feature maps in the deeper layers. The number of maps in the large network is summarized in Table \ref{tab:layers}.
We generalize the initialization scheme to account for the difference in the number of feature maps in different layers. Let $c_i$ and $c_{i+1}$ be the number of feature maps in two consecutive layers. Assume that $C$ divides both $c_i$ and $c_{i+1}$. The initialization is
\begin{equation}
  k^b (\tt, a) = \begin{cases}
  \displaystyle {C \over c_{i+1}} \quad & \displaystyle \tt = 0 \textup{\ \ and\ \ } \left\lfloor{a C \over c_i}\right\rfloor = \left\lfloor{b C \over c_{i+1}}\right\rfloor \vspace{2mm}\\
  \quad \!\varepsilon \quad & \displaystyle \textup{otherwise}
  \end{cases}
\end{equation}

Here $\varepsilon \sim \N(0,\sigma^2)$ and $\sigma \ll C / c_{i+1}$.
The use of random noise breaks ties among feature maps with a common predecessor.

\begin{figure}[t]
    \centering
  \begin{subfigure}[b]{0.19\linewidth}
    \includegraphics[width=\textwidth]{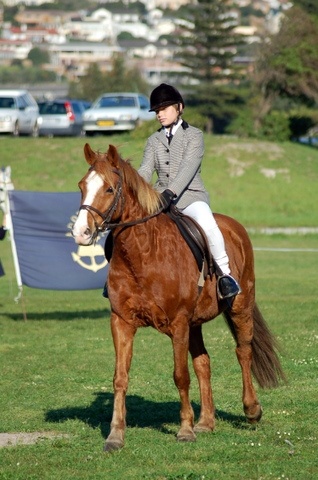}
  \end{subfigure}
  \begin{subfigure}[b]{0.19\linewidth}
    \includegraphics[width=\textwidth]{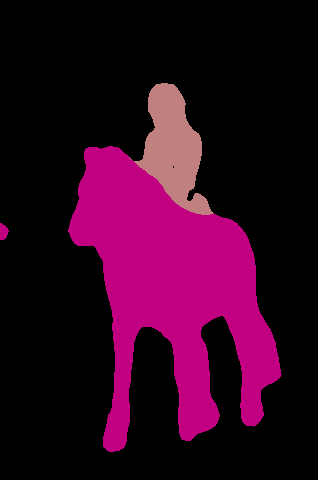}
  \end{subfigure}
  \begin{subfigure}[b]{0.19\linewidth}
    \includegraphics[width=\textwidth]{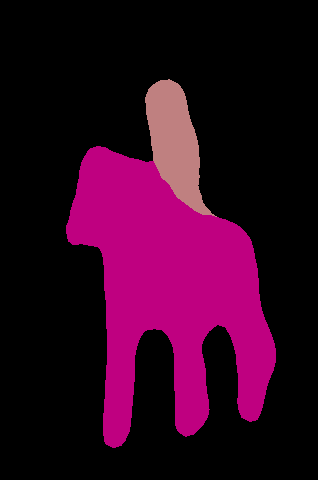}
  \end{subfigure}
  \begin{subfigure}[b]{0.19\linewidth}
    \includegraphics[width=\textwidth]{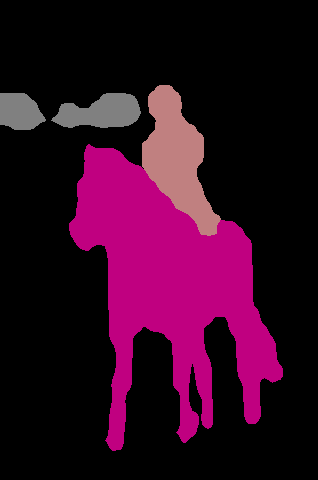}
  \end{subfigure}
  \begin{subfigure}[b]{0.19\linewidth}
    \includegraphics[width=\textwidth]{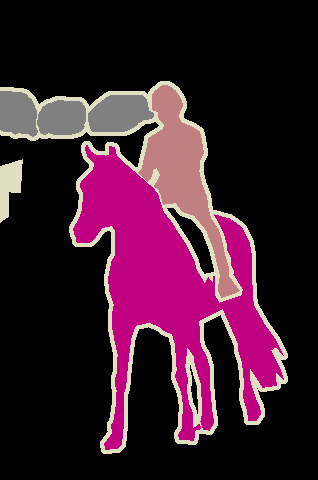}
  \end{subfigure}

  \begin{subfigure}[b]{0.19\linewidth}
    \includegraphics[width=\textwidth]{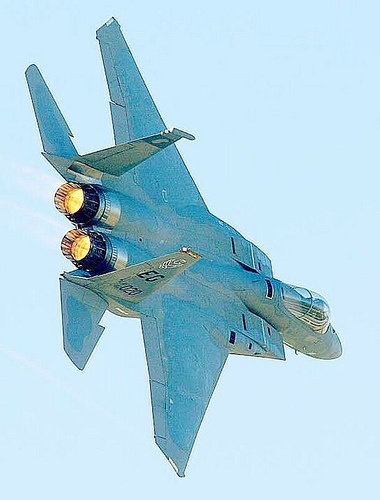}
  \end{subfigure}
  \begin{subfigure}[b]{0.19\linewidth}
    \includegraphics[width=\textwidth]{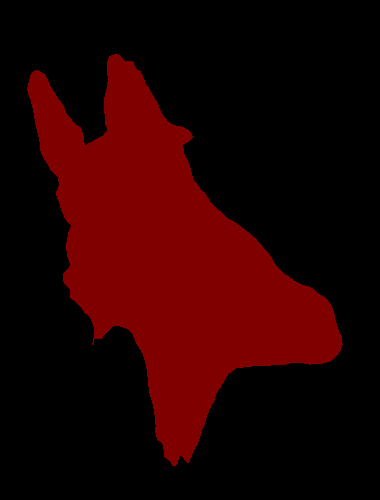}
  \end{subfigure}
  \begin{subfigure}[b]{0.19\linewidth}
    \includegraphics[width=\textwidth]{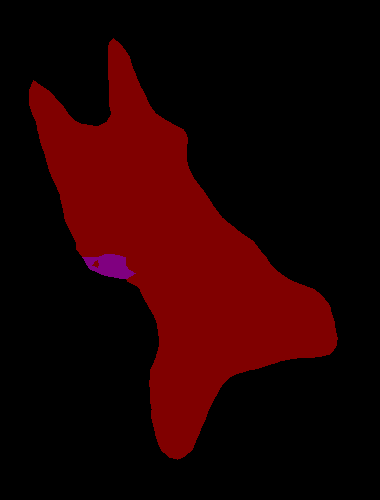}
  \end{subfigure}
  \begin{subfigure}[b]{0.19\linewidth}
    \includegraphics[width=\textwidth]{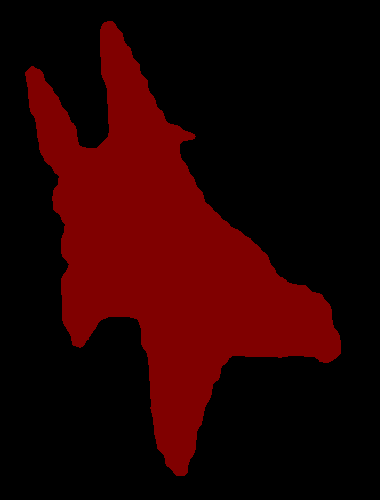}
  \end{subfigure}
  \begin{subfigure}[b]{0.19\linewidth}
    \includegraphics[width=\textwidth]{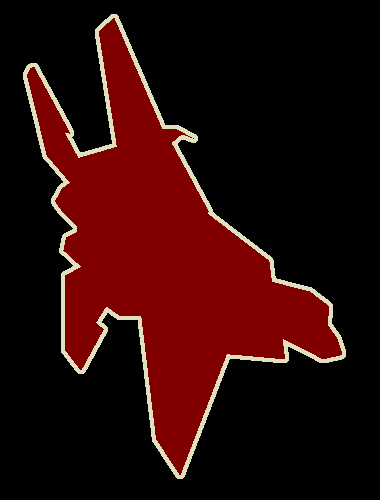}
  \end{subfigure}

  \begin{subfigure}[b]{0.19\linewidth}
    \includegraphics[width=\textwidth]{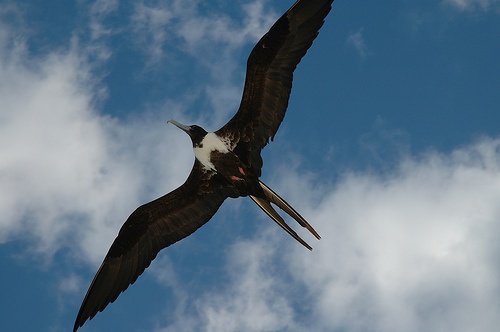}
  \end{subfigure}
  \begin{subfigure}[b]{0.19\linewidth}
    \includegraphics[width=\textwidth]{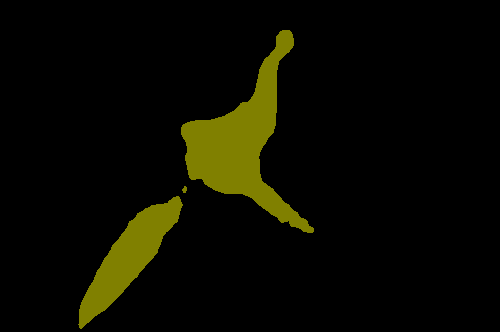}
  \end{subfigure}
  \begin{subfigure}[b]{0.19\linewidth}
    \includegraphics[width=\textwidth]{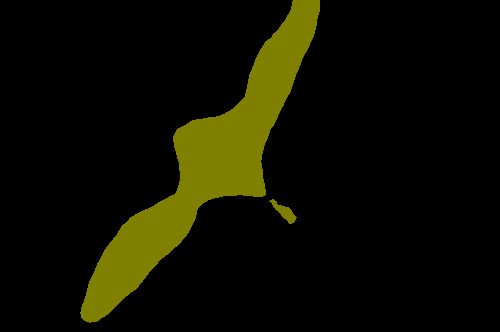}
  \end{subfigure}
  \begin{subfigure}[b]{0.19\linewidth}
    \includegraphics[width=\textwidth]{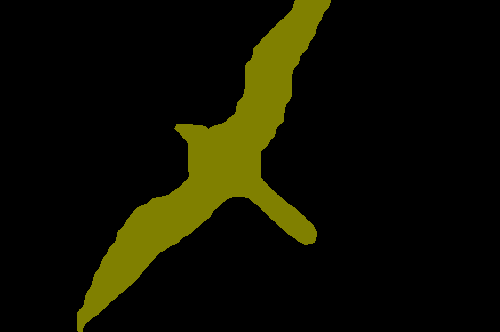}
  \end{subfigure}
  \begin{subfigure}[b]{0.19\linewidth}
    \includegraphics[width=\textwidth]{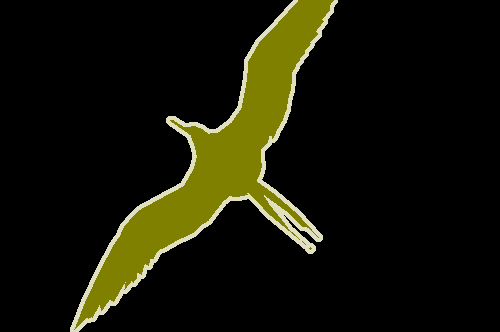}
  \end{subfigure}

  \begin{subfigure}[b]{0.19\linewidth}
    \includegraphics[width=\textwidth]{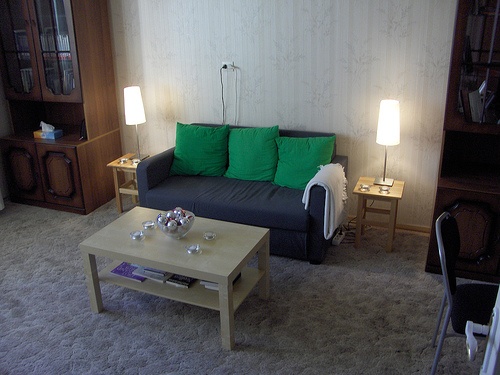}
  \end{subfigure}
  \begin{subfigure}[b]{0.19\linewidth}
    \includegraphics[width=\textwidth]{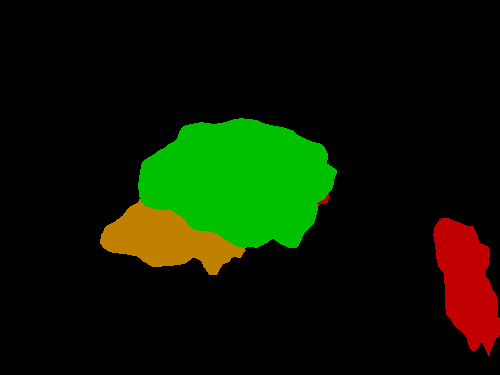}
  \end{subfigure}
  \begin{subfigure}[b]{0.19\linewidth}
    \includegraphics[width=\textwidth]{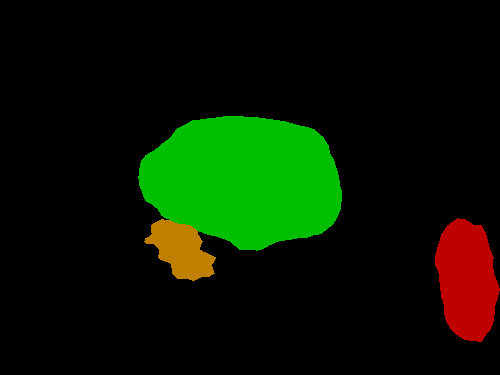}
  \end{subfigure}
  \begin{subfigure}[b]{0.19\linewidth}
    \includegraphics[width=\textwidth]{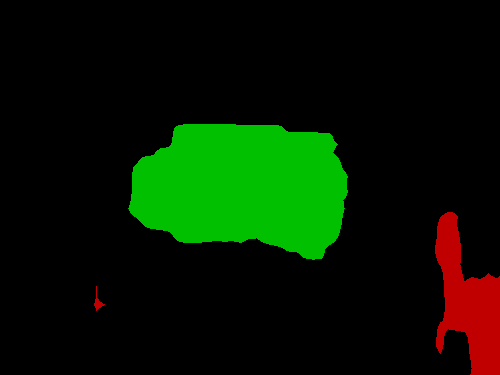}
  \end{subfigure}
  \begin{subfigure}[b]{0.19\linewidth}
    \includegraphics[width=\textwidth]{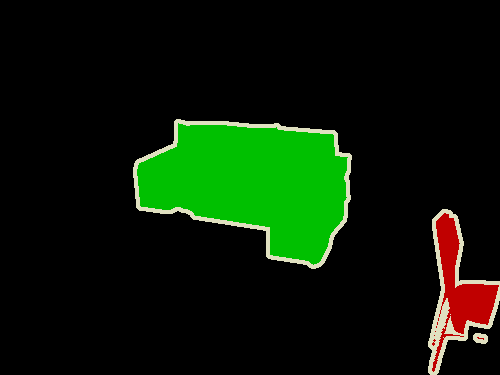}
  \end{subfigure}

  \begin{subfigure}[b]{0.19\linewidth}
    \includegraphics[width=\textwidth]{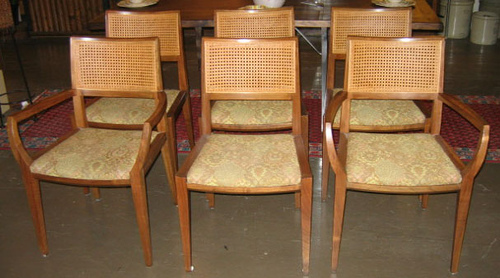}
        \caption{Image}
  \end{subfigure}
  \begin{subfigure}[b]{0.19\linewidth}
    \includegraphics[width=\textwidth]{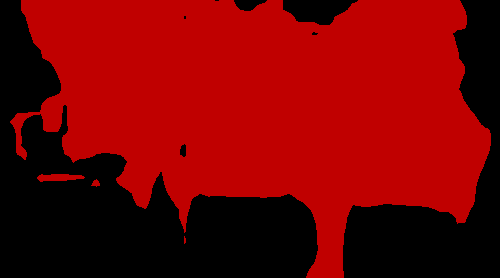}
        \caption{FCN-8s}
  \end{subfigure}
  \begin{subfigure}[b]{0.19\linewidth}
    \includegraphics[width=\textwidth]{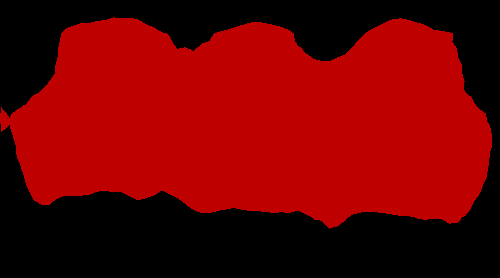}
        \caption{DeepLab}
  \end{subfigure}
  \begin{subfigure}[b]{0.19\linewidth}
    \includegraphics[width=\textwidth]{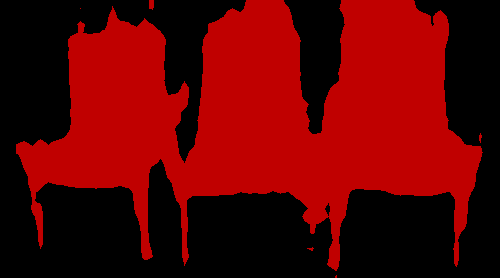}
        \caption{Our front end}
  \end{subfigure}
  \begin{subfigure}[b]{0.19\linewidth}
    \includegraphics[width=\textwidth]{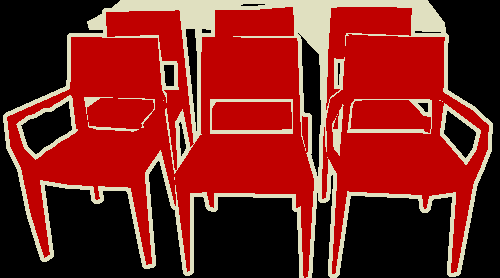}
        \caption{Ground truth}
  \end{subfigure}

\vspace{-2mm}
\caption{Semantic segmentations produced by different adaptations of the VGG-16 classification network. From left to right: (a) input image, (b) prediction by FCN-8s \citep{Long2015}, (c) prediction by DeepLab \citep{Chen2015ICLR}, (d) prediction by our simplified front-end module, (e) ground truth.}
\label{fig:unary}

\vspace{2mm}

\begingroup
\setlength{\tabcolsep}{1.5pt}
\scriptsize
\begin{center}
\begin{tabular}{l||c|c|c|c|c|c|c|c|c|c|c|c|c|c|c|c|c|c|c|c||c}
 & \ver{aero} & \ver{bike} & \ver{bird} & \ver{boat} & \ver{bottle} &
  \ver{bus} & \ver{car} & \ver{cat} & \ver{chair} & \ver{cow} &
  \ver{table} & \ver{dog} & \ver{horse} & \ver{mbike} & \ver{person} &
  \ver{plant} & \ver{sheep} & \ver{sofa} & \ver{train} & \ver{tv} &
  \ver{\,mean IoU} \\ \hline
FCN-8s & 76.8 & 34.2 & 68.9 & 49.4 & 60.3 & 75.3 & 74.7 & 77.6 & 21.4
& 62.5 & 46.8 & 71.8 & 63.9 & 76.5 & 73.9 & 45.2 & 72.4 & 37.4 & 70.9
& 55.1 & 62.2 \\
DeepLab & 72 & 31 & 71.2 & 53.7 & 60.5 & 77 & 71.9 & 73.1 & 25.2 &
62.6 & 49.1 & 68.7 & 63.3 & 73.9 & 73.6 & 50.8 & 72.3 & 42.1 & 67.9 &
52.6 & 62.1 \\
DeepLab-Msc & 74.9 & 34.1 & 72.6 & 52.9 & 61.0 & 77.9 & 73.0 & 73.7 &
26.4 & 62.2 & 49.3 & 68.4 & 64.1 & 74.0 & 75.0 & 51.7 & 72.7 & 42.5 &
67.2 & 55.7 & 62.9 \\
Our front end & \textbf{82.2} & \textbf{37.4} & \textbf{72.7} &
\textbf{57.1} & \textbf{62.7} & \textbf{82.8} & \textbf{77.8} &
\textbf{78.9} & \textbf{28} & \textbf{70} & \textbf{51.6} &
\textbf{73.1} & \textbf{72.8} & \textbf{81.5} & \textbf{79.1} &
\textbf{56.6} & \textbf{77.1} & \textbf{49.9} & \textbf{75.3} &
\textbf{60.9} & \textbf{67.6} \\ \hline
\end{tabular}
\end{center}
\endgroup
\vspace{-2mm}
\captionof{table}{Our front-end prediction module is simpler and more accurate than prior models. This table reports accuracy on the VOC-2012 test set.}
\label{tab:unary}

\vspace{-5mm}

\end{figure}

\section{Front End}
\label{sec:front-end}

We implemented and trained a front-end prediction module that takes a color image as input and produces $C=21$ feature maps as output. The front-end module follows the work of \cite{Long2015} and \cite{Chen2015ICLR}, but was implemented separately. We adapted the VGG-16 network \citep{SimonyanZisserman2015} for dense prediction and removed the last two pooling and striding layers. Specifically, each of these pooling and striding layers was removed and convolutions in all subsequent layers were dilated by a factor of 2 for each pooling layer that was ablated. Thus convolutions in the final layers, which follow both ablated pooling layers, are dilated by a factor of 4. This enables initialization with the parameters of the original classification network, but produces higher-resolution output. The front-end module takes padded images as input and produces feature maps at resolution $64\timess 64$. We use reflection padding: the buffer zone is filled by reflecting the image about each edge.

Our front-end module is obtained by removing vestiges of the classification network that are counter-productive for dense prediction. Most significantly, we remove the last two pooling and striding layers entirely, whereas Long et al.~kept them and Chen et al.~replaced striding by dilation but kept the pooling layers. We found that simplifying the network by removing the pooling layers made it more accurate. We also remove the padding of the intermediate feature maps. Intermediate padding was used in the original classification network, but is neither necessary nor justified in dense prediction.

This simplified prediction module was trained on the Pascal VOC 2012 training set, augmented by the annotations created by \cite{Hariharan2011}. We did not use images from the VOC-2012 validation set for training and therefore only used a subset of the annotations of \cite{Hariharan2011}. Training was performed by stochastic gradient descent (SGD) with mini-batch size 14, learning rate $10^{-3}$, and momentum $0.9$. The network was trained for 60K iterations.

We now compare the accuracy of our front-end module to the FCN-8s design of \cite{Long2015} and the DeepLab network of \cite{Chen2015ICLR}. For FCN-8s and DeepLab, we evaluate the public models trained by the original authors on VOC-2012. Segmentations produced by the different models on images from the VOC-2012 dataset are shown in Figure \ref{fig:unary}. The accuracy of the models on the VOC-2012 test set is reported in Table \ref{tab:unary}.

Our front-end prediction module is both simpler and more accurate than the prior models. Specifically, our simplified model outperforms both FCN-8s and the DeepLab network by more than 5 percentage points on the test set. Interestingly, our simplified front-end module outperforms the leaderboard accuracy of DeepLab+CRF on the test set by more than a percentage point ($67.6\%$ vs.~$66.4\%$) without using a CRF.

\section{Experiments}
\label{sec:evaluation}

Our implementation is based on the Caffe library \citep{Jia2014}. Our implementation of dilated convolutions is now part of the stanfard Caffe distribution.

For fair comparison with recent high-performing systems, we trained a front-end module that has the same structure as described in Section \ref{sec:front-end}, but is trained on additional images from the Microsoft COCO dataset~\citep{Lin2014}. We used all images in Microsoft COCO with at least one object from the VOC-2012 categories. Annotated objects from other categories were treated as background.

Training was performed in two stages. In the first stage, we trained on VOC-2012 images and Microsoft COCO images together. Training was performed by SGD with mini-batch size 14 and momentum 0.9. 100K iterations were performed with a learning rate of $10^{-3}$ and 40K subsequent iterations were performed with a learning rate of $10^{-4}$. In the second stage, we fine-tuned the network on \mbox{VOC-2012} images only. Fine-tuning was performed for 50K iterations with a learning rate of $10^{-5}$. Images from the VOC-2012 validation set were not used for training.

The front-end module trained by this procedure achieves $69.8\%$ mean IoU on the VOC-2012 validation set and $71.3\%$ mean IoU on the test set. Note that this level of accuracy is achieved by the front-end alone, without the context module or structured prediction. We again attribute this high accuracy in part to the removal of vestigial components originally developed for image classification rather than dense prediction.

\paragraph{Controlled evaluation of context aggregation.}
We now perform controlled experiments to evaluate the utility of the context network presented in Section \ref{sec:context}. We begin by plugging each of the two context modules (Basic and Large) into the front end. Since the receptive field of the context network is $67\timess 67$, we pad the input feature maps by a buffer of width 33. Zero padding and reflection padding yielded similar results in our experiments. The context module accepts feature maps from the front end as input and is given this input during training.
Joint training of the context module and the front-end module did not yield a significant improvement in our experiments. The learning rate was set to $10^{-3}$. Training was initialized as described in Section \ref{sec:context}.

\begin{figure}[t]
    \centering
  \begin{subfigure}[b]{0.19\linewidth}
    \includegraphics[width=\textwidth]{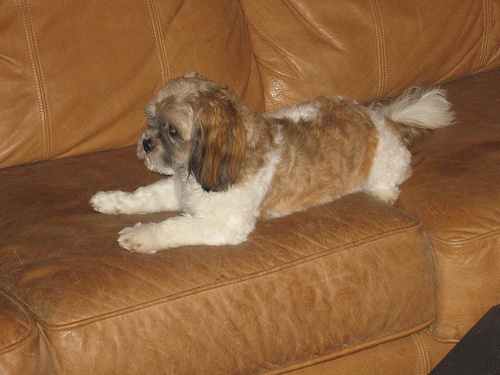}
  \end{subfigure}
  \begin{subfigure}[b]{0.19\linewidth}
    \includegraphics[width=\textwidth]{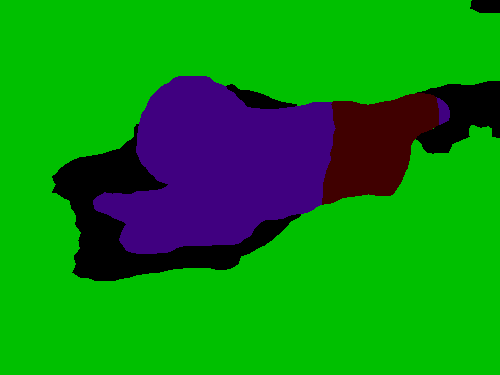}
  \end{subfigure}
  \begin{subfigure}[b]{0.19\linewidth}
    \includegraphics[width=\textwidth]{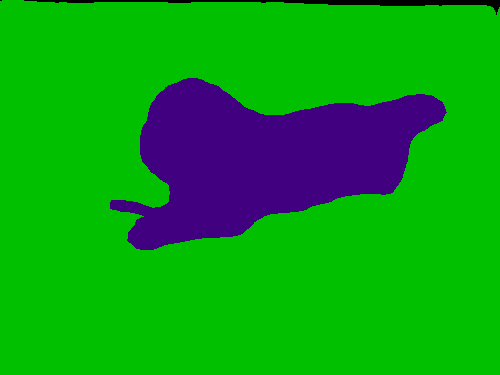}
  \end{subfigure}
  \begin{subfigure}[b]{0.19\linewidth}
    \includegraphics[width=\textwidth]{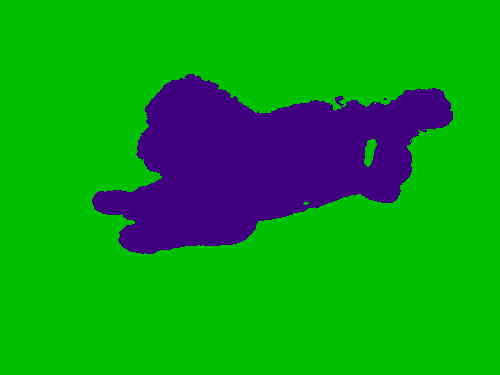}
  \end{subfigure}
  \begin{subfigure}[b]{0.19\linewidth}
    \includegraphics[width=\textwidth]{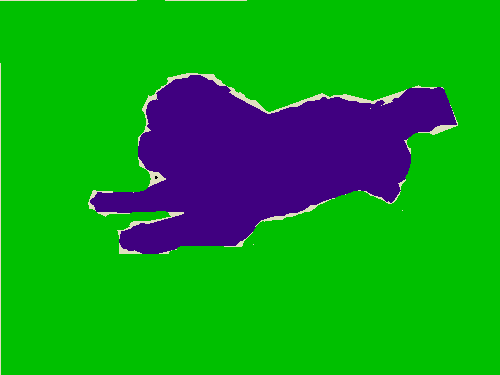}
  \end{subfigure}

  \begin{subfigure}[b]{0.19\linewidth}
    \includegraphics[width=\textwidth]{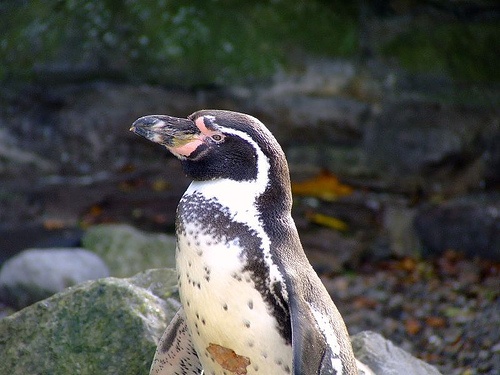}
  \end{subfigure}
  \begin{subfigure}[b]{0.19\linewidth}
    \includegraphics[width=\textwidth]{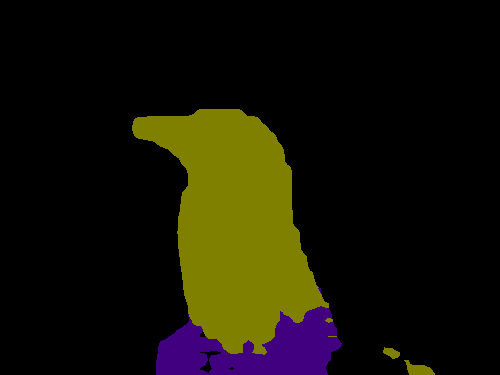}
  \end{subfigure}
  \begin{subfigure}[b]{0.19\linewidth}
    \includegraphics[width=\textwidth]{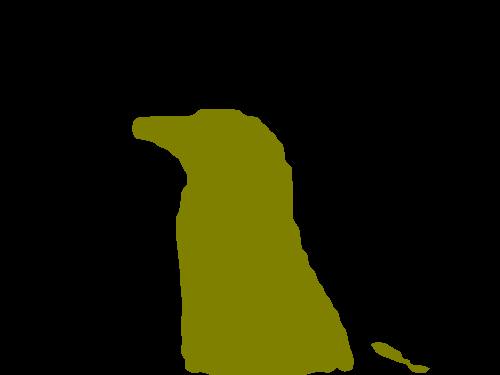}
  \end{subfigure}
  \begin{subfigure}[b]{0.19\linewidth}
    \includegraphics[width=\textwidth]{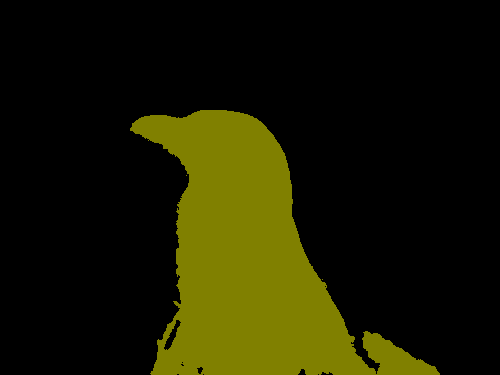}
  \end{subfigure}
  \begin{subfigure}[b]{0.19\linewidth}
    \includegraphics[width=\textwidth]{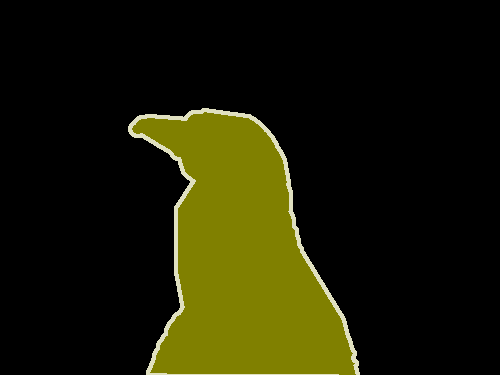}
  \end{subfigure}

  \begin{subfigure}[b]{0.19\linewidth}
    \includegraphics[width=\textwidth]{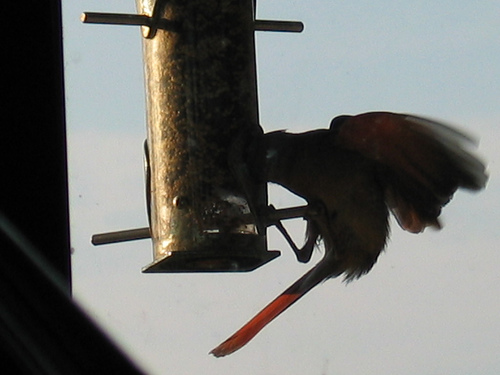}
  \end{subfigure}
  \begin{subfigure}[b]{0.19\linewidth}
    \includegraphics[width=\textwidth]{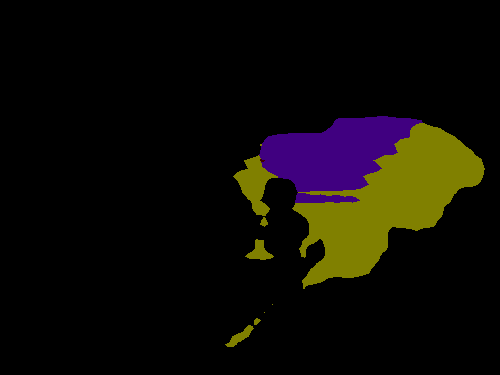}
  \end{subfigure}
  \begin{subfigure}[b]{0.19\linewidth}
    \includegraphics[width=\textwidth]{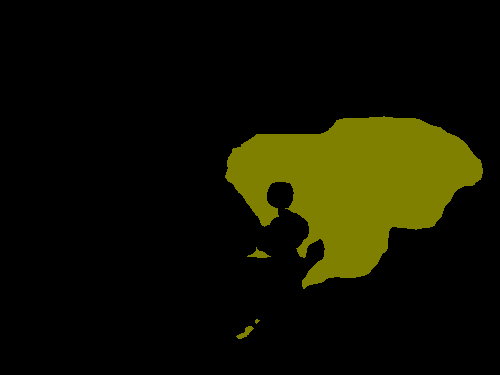}
  \end{subfigure}
  \begin{subfigure}[b]{0.19\linewidth}
    \includegraphics[width=\textwidth]{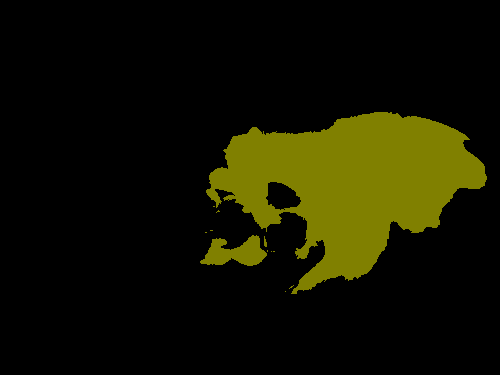}
  \end{subfigure}
  \begin{subfigure}[b]{0.19\linewidth}
    \includegraphics[width=\textwidth]{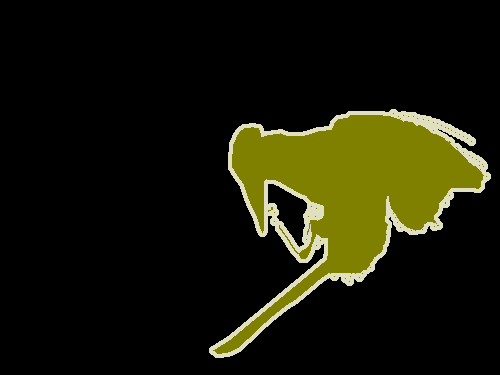}
  \end{subfigure}

  \begin{subfigure}[b]{0.19\linewidth}
    \includegraphics[width=\textwidth]{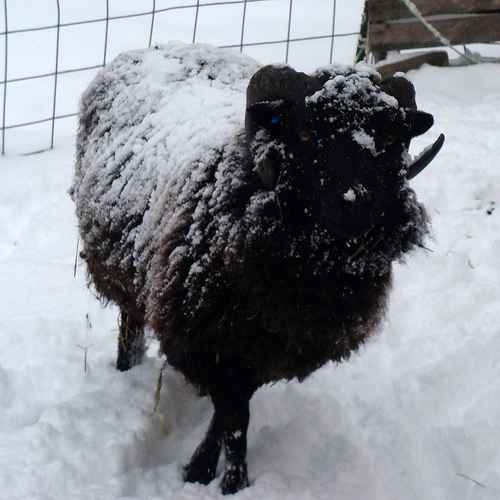}
  \end{subfigure}
  \begin{subfigure}[b]{0.19\linewidth}
    \includegraphics[width=\textwidth]{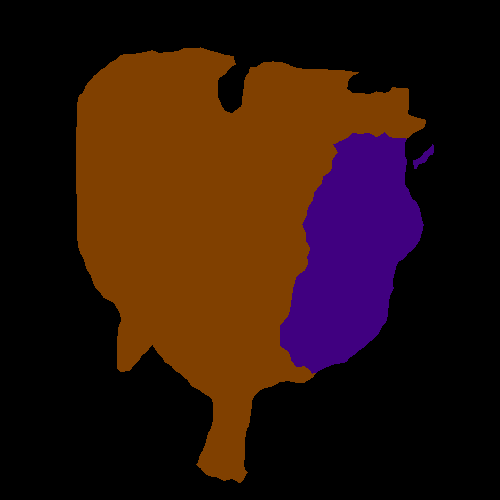}
  \end{subfigure}
  \begin{subfigure}[b]{0.19\linewidth}
    \includegraphics[width=\textwidth]{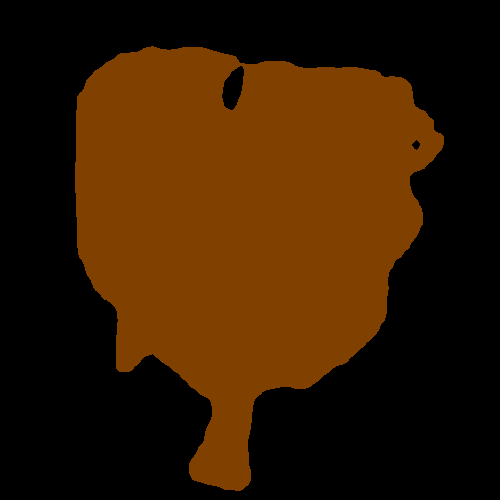}
  \end{subfigure}
  \begin{subfigure}[b]{0.19\linewidth}
    \includegraphics[width=\textwidth]{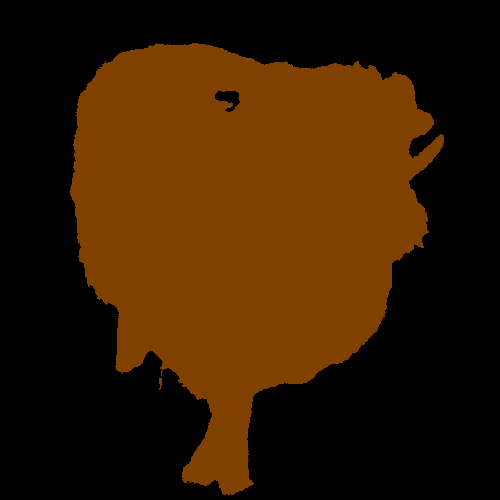}
  \end{subfigure}
  \begin{subfigure}[b]{0.19\linewidth}
    \includegraphics[width=\textwidth]{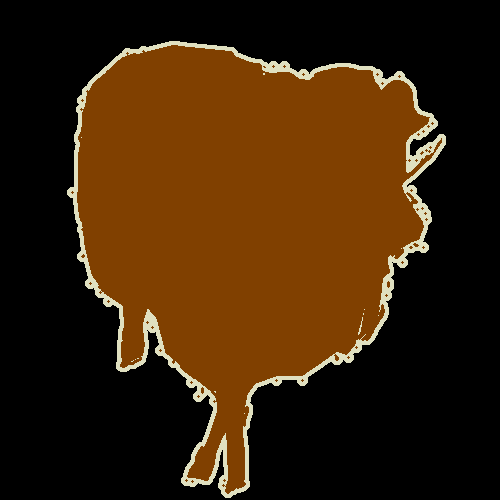}
  \end{subfigure}

  \begin{subfigure}[b]{0.19\linewidth}
    \includegraphics[width=\textwidth]{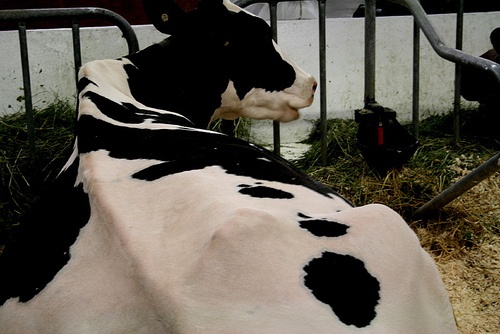}
	\caption{Image}
  \end{subfigure}
  \begin{subfigure}[b]{0.19\linewidth}
    \includegraphics[width=\textwidth]{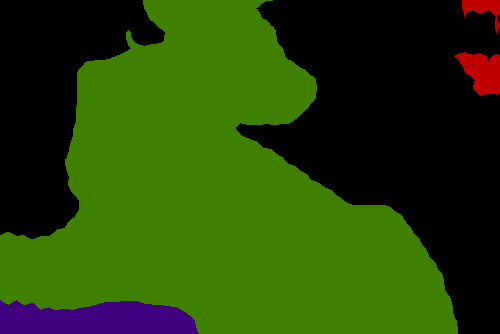}
	\caption{Front end}
  \end{subfigure}
  \begin{subfigure}[b]{0.19\linewidth}
    \includegraphics[width=\textwidth]{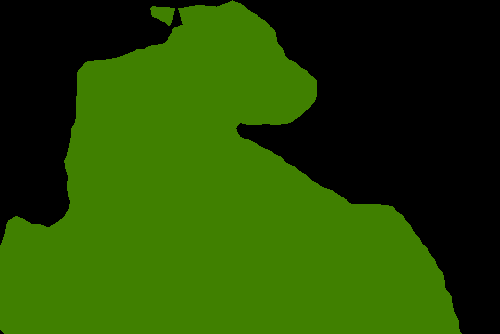}
	\caption{\,+\,Context}
  \end{subfigure}
  \begin{subfigure}[b]{0.19\linewidth}
    \includegraphics[width=\textwidth]{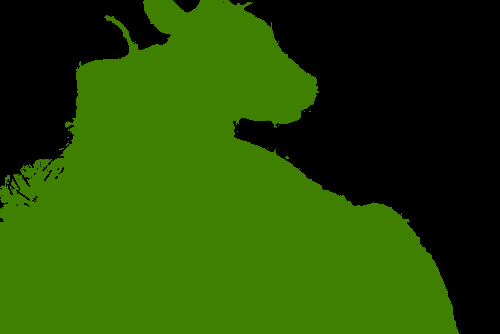}
	\caption{\,+\,CRF-RNN}
  \end{subfigure}
  \begin{subfigure}[b]{0.19\linewidth}
    \includegraphics[width=\textwidth]{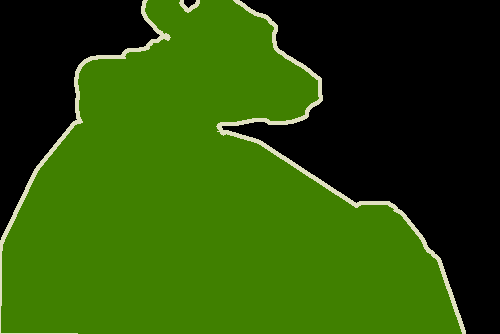}
	\caption{Ground truth}
  \end{subfigure}

\vspace{-2mm}
\caption{Semantic segmentations produced by different models. From left to right: (a) input image, (b) prediction by the front-end module, (c) prediction by the large context network plugged into the front end, (d) prediction by the front end + context module + CRF-RNN, (e) ground truth.}
\label{fig:controlled}

\vspace{2mm}

\begingroup
\setlength{\tabcolsep}{1.5pt}
\scriptsize
\begin{center}
\begin{tabular}{l||c|c|c|c|c|c|c|c|c|c|c|c|c|c|c|c|c|c|c|c||c}
 & \ver{aero} & \ver{bike} & \ver{bird} & \ver{boat} & \ver{bottle} &
  \ver{bus} & \ver{car} & \ver{cat} & \ver{chair} & \ver{cow} &
  \ver{table} & \ver{dog} & \ver{horse} & \ver{mbike} & \ver{person} &
  \ver{plant} & \ver{sheep} & \ver{sofa} & \ver{train} & \ver{tv} &
  \ver{\,mean IoU} \\ \hline
Front end & 86.3 & 38.2 & 76.8 & \textbf{66.8} & 63.2 & 87.3 & 78.7 &
82 & 33.7 & 76.7 & 53.5 & 73.7 & 76 & 76.6 & 83 & \textbf{51.9} & 77.8
& 44 & 79.9 & \textbf{66.3} & 69.8 \\
Front + Basic & 86.4 & 37.6 & 78.5 & 66.3 & 64.1 & 89.9 & 79.9 & 84.9
& \textbf{36.1} & 79.4 & \textbf{55.8} & 77.6 & 81.6 & 79 & 83.1 &
51.2 & 81.3 & 43.7 & 82.3 & 65.7 & 71.3 \\
Front + Large & \textbf{87.3} & \textbf{39.2} & \textbf{80.3} & 65.6 &
\textbf{66.4} & \textbf{90.2} & \textbf{82.6} & \textbf{85.8} & 34.8 &
\textbf{81.9} & 51.7 & \textbf{79} & \textbf{84.1} & \textbf{80.9} &
\textbf{83.2} & 51.2 & \textbf{83.2} & \textbf{44.7} & \textbf{83.4} &
65.6 & \textbf{72.1} \\ \hline
Front end + CRF & 89.2 & 38.8 & 80 & \textbf{69.8} & 63.2 & 88.8 & 80
& 85.2 & 33.8 & 80.6 & 55.5 & 77.1 & 80.8 & 77.3 & 84.3 &
\textbf{53.1} & 80.4 & 45 & 80.7 & \textbf{67.9} & 71.6 \\
Front + Basic + CRF & 89.1 & 38.7 & 81.4 & 67.4 & 65 & 91 & 81 & 86.7
& \textbf{37.5} & 81 & \textbf{57} & 79.6 & 83.6 & 79.9 &
\textbf{84.6} & 52.7 & 83.3 & 44.3 & 82.6 & 67.2 & 72.7 \\
Front + Large + CRF & \textbf{89.6} & \textbf{39.9} & \textbf{82.7} &
66.7 & \textbf{67.5} & \textbf{91.1} & \textbf{83.3} & \textbf{87.4} &
36 & \textbf{83.3} & 52.5 & \textbf{80.7} & \textbf{85.7} &
\textbf{81.8} & 84.4 & 52.6 & \textbf{84.4} & \textbf{45.3} &
\textbf{83.7} & 66.7 & \textbf{73.3} \\ \hline
Front end + RNN & 88.8 & 38.1 & 80.8 & \textbf{69.1} & 65.6 & 89.9 &
79.6 & 85.7 & 36.3 & 83.6 & 57.3 & 77.9 & 83.2 & 77 & \textbf{84.6} &
\textbf{54.7} & 82.1 & \textbf{46.9} & 80.9 & 66.7 & 72.5 \\
Front + Basic + RNN & 89 & 38.4 & 82.3 & 67.9 & 65.2 & 91.5 & 80.4 &
87.2 & \textbf{38.4} & 82.1 & \textbf{57.7} & 79.9 & 85 & 79.6 & 84.5
& 53.5 & 84 & 45 & 82.8 & 66.2 & 73.1 \\
Front + Large + RNN & \textbf{89.3} & \textbf{39.2} & \textbf{83.6} &
67.2 & \textbf{69} & \textbf{92.1} & \textbf{83.1} & \textbf{88} &
\textbf{38.4} & \textbf{84.8} & 55.3 & \textbf{81.2} & \textbf{86.7} &
\textbf{81.3} & 84.3 & 53.6 & \textbf{84.4} & 45.8 & \textbf{83.8} &
\textbf{67} & \textbf{73.9} \\ \hline
\end{tabular}
\end{center}
\endgroup
\vspace{-2mm}
\captionof{table}{Controlled evaluation of the effect of the context module on the accuracy of three different architectures for semantic segmentation. Experiments performed on the VOC-2012 validation set. Validation images were not used for training. Top: adding the context module to a semantic segmentation front end with no structured prediction \citep{Long2015}. The basic context module increases accuracy, the large module increases it by a larger margin. Middle: the context module increases accuracy when plugged into a front-end + dense CRF configuration \citep{Chen2015ICLR}. Bottom: the context module increases accuracy when plugged into a front-end + CRF-RNN configuration \citep{Zheng2015}.}
\label{tab:controlled}

\vspace{-3mm}

\end{figure}

Table \ref{tab:controlled} shows the effect of adding the context module to three different architectures for semantic segmentation. The first architecture (top) is the front end described in Section \ref{sec:front-end}. It performs semantic segmentation without structured prediction, akin to the original work of \cite{Long2015}. The second architecture (Table \ref{tab:controlled}, middle) uses the dense CRF to perform structured prediction, akin to the system of \cite{Chen2015ICLR}. We use the implementation of \cite{KrahenbuhlKoltun2011} and train the CRF parameters by grid search on the validation set. The third architecture (Table \ref{tab:controlled}, bottom) uses the CRF-RNN for structured prediction \citep{Zheng2015}. We use the implementation of \cite{Zheng2015} and train the CRF-RNN in each condition.

The experimental results demonstrate that the context module improves accuracy in each of the three configurations. The basic context module increases accuracy in each configuration. The large context module increases accuracy by a larger margin. The experiments indicate that the context module and structured prediction are synergisic: the context module increases accuracy with or without subsequent structured prediction. Qualitative results are shown in Figure \ref{fig:controlled}.

\paragraph{Evaluation on the test set.}
We now perform an evaluation on the test set by submitting our results to the Pascal VOC 2012 evaluation server. The results are reported in Table \ref{tab:test}. We use the large context module for these experiments. As the results demonstrate, the context module yields a significant boost in accuracy over the front end. The context module alone, without subsequent structured prediction, outperforms DeepLab-CRF-COCO-LargeFOV \citep{Chen2015ICLR}. The context module with the dense CRF, using the original implementation of \cite{KrahenbuhlKoltun2011}, performs on par with the very recent CRF-RNN \citep{Zheng2015}. The context module in combination with the CRF-RNN further increases accuracy over the performance of the CRF-RNN.


\begin{table*}[t]
\begingroup
\setlength{\tabcolsep}{1.5pt}
\scriptsize
\begin{center}
\begin{tabular}{l||c|c|c|c|c|c|c|c|c|c|c|c|c|c|c|c|c|c|c|c||c}
 & \ver{aero} & \ver{bike} & \ver{bird} & \ver{boat} & \ver{bottle} &
  \ver{bus} & \ver{car} & \ver{cat} & \ver{chair} & \ver{cow} &
  \ver{table} & \ver{dog} & \ver{horse} & \ver{mbike} & \ver{person} &
  \ver{plant} & \ver{sheep} & \ver{sofa} & \ver{train} & \ver{tv} &
  \ver{\,mean IoU} \\ \hline
DeepLab++ & 89.1 & 38.3 & 88.1 & 63.3 & 69.7 & 87.1 & \textbf{83.1} &
85 & 29.3 & 76.5 & 56.5 & 79.8 & 77.9 & 85.8 & 82.4 & 57.4 & 84.3 &
54.9 & 80.5 & 64.1 & 72.7 \\
DeepLab-MSc++  & 89.2 & 46.7 & 88.5 & 63.5 & 68.4 & 87.0 & 81.2 & 86.3
& 32.6 & 80.7 & 62.4 & 81.0 & 81.3 & 84.3 & 82.1 & 56.2 & 84.6 & 58.3
& 76.2 & 67.2 & 73.9 \\
CRF-RNN & 90.4 & \textbf{55.3} & 88.7 & \textbf{68.4} & 69.8 & 88.3 &
82.4 & 85.1 & 32.6 & 78.5 & \textbf{64.4} & 79.6 & 81.9 &
\textbf{86.4} & 81.8 & \textbf{58.6} & 82.4 & 53.5 & 77.4 &
\textbf{70.1} & 74.7 \\ \hline
Front end & 86.6 & 37.3 & 84.9 & 62.4 & 67.3 & 86.2 & 81.2 & 82.1 &
32.6 & 77.4 & 58.3 & 75.9 & 81 & 83.6 & 82.3 & 54.2 & 81.5 & 50.1 &
77.5 & 63 & 71.3 \\
Context & 89.1 & 39.1 & 86.8 & 62.6 & 68.9 & 88.2 & 82.6 & 87.7 & 33.8
& 81.2 & 59.2 & 81.8 & 87.2 & 83.3 & 83.6 & 53.6 & 84.9 & 53.7 & 80.5
& 62.9 & 73.5 \\
Context + CRF & 91.3 & 39.9 & \textbf{88.9} & 64.3 & 69.8 & 88.9 &
82.6 & 89.7 & 34.7 & 82.7 & 59.5 & 83 & 88.4 & 84.2 & 85 & 55.3 & 86.7
& 54.4 & \textbf{81.9} & 63.6 & 74.7 \\
Context + CRF-RNN  & \textbf{91.7} & 39.6 & 87.8 & 63.1 &
\textbf{71.8} & \textbf{89.7} & 82.9 & \textbf{89.8} & \textbf{37.2} &
\textbf{84} & 63 & \textbf{83.3} & \textbf{89} & 83.8 & \textbf{85.1}
& 56.8 & \textbf{87.6} & \textbf{56} & 80.2 & 64.7 & \textbf{75.3} \\ \hline
\end{tabular}
\vspace{-2mm}
\end{center}
\endgroup
\caption{Evaluation on the VOC-2012 test set. `DeepLab++' stands for DeepLab-CRF-COCO-LargeFOV and `DeepLab-MSc++' stands for DeepLab-MSc-CRF-LargeFOV-COCO-CrossJoint \citep{Chen2015ICLR}. `CRF-RNN' is the system of \cite{Zheng2015}. `Context' refers to the large context module plugged into our front end. The context network yields very high accuracy, ourperforming the DeepLab++ architecture without performing structured prediction. Combining the context network with the CRF-RNN structured prediction module increases the accuracy of the CRF-RNN system.}
\label{tab:test}
\end{table*}

\section{Conclusion}

We have examined convolutional network architectures for dense prediction. Since the model must produce high-resolution output, we believe that high-resolution operation throughout the network is both feasible and desirable. Our work shows that the dilated convolution operator is particularly suited to dense prediction due to its ability to expand the receptive field without losing resolution or coverage. We have utilized dilated convolutions to design a new network structure that reliably increases accuracy when plugged into existing semantic segmentation systems. As part of this work, we have also shown that the accuracy of existing convolutional networks for semantic segmentation can be increased by removing vestigial components that had been developed for image classification.

We believe that the presented work is a step towards dedicated architectures for dense prediction that are not constrained by image classification precursors. As new sources of data become available, future architectures may be trained densely end-to-end, removing the need for pre-training on image classification datasets. This may enable architectural simplification and unification. Specifically, end-to-end dense training may enable a fully dense architecture akin to the presented context network to operate at full resolution throughout, accepting the raw image as input and producing dense label assignments at full resolution as output.

\begin{figure}[b]
  \centering
  \begin{subfigure}[b]{0.15\linewidth}
    \includegraphics[width=\textwidth]{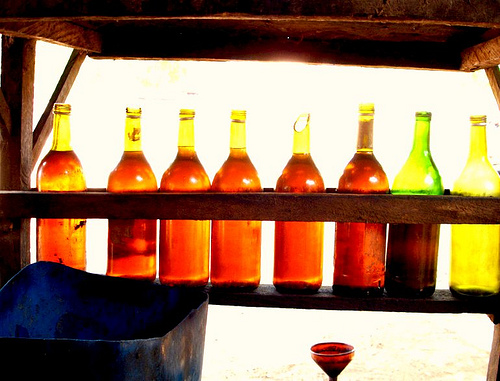}
  \end{subfigure}
  \begin{subfigure}[b]{0.15\linewidth}
    \includegraphics[width=\textwidth]{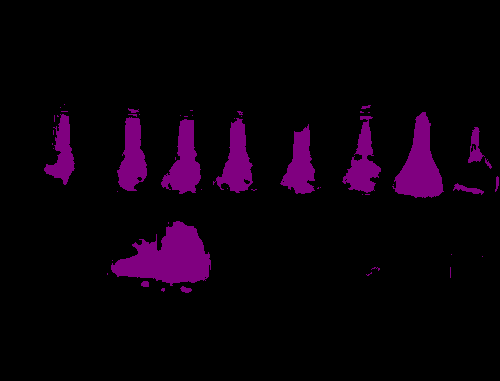}
  \end{subfigure}
  \begin{subfigure}[b]{0.15\linewidth}
    \includegraphics[width=\textwidth]{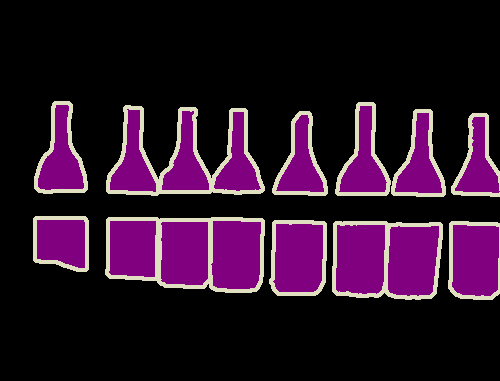}
  \end{subfigure}
  ~
  \begin{subfigure}[b]{0.15\linewidth}
    \includegraphics[width=\textwidth]{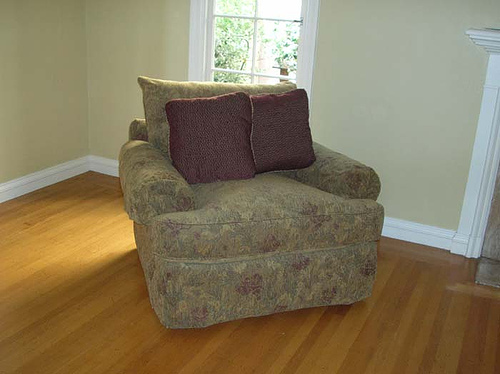}
  \end{subfigure}
  \begin{subfigure}[b]{0.15\linewidth}
    \includegraphics[width=\textwidth]{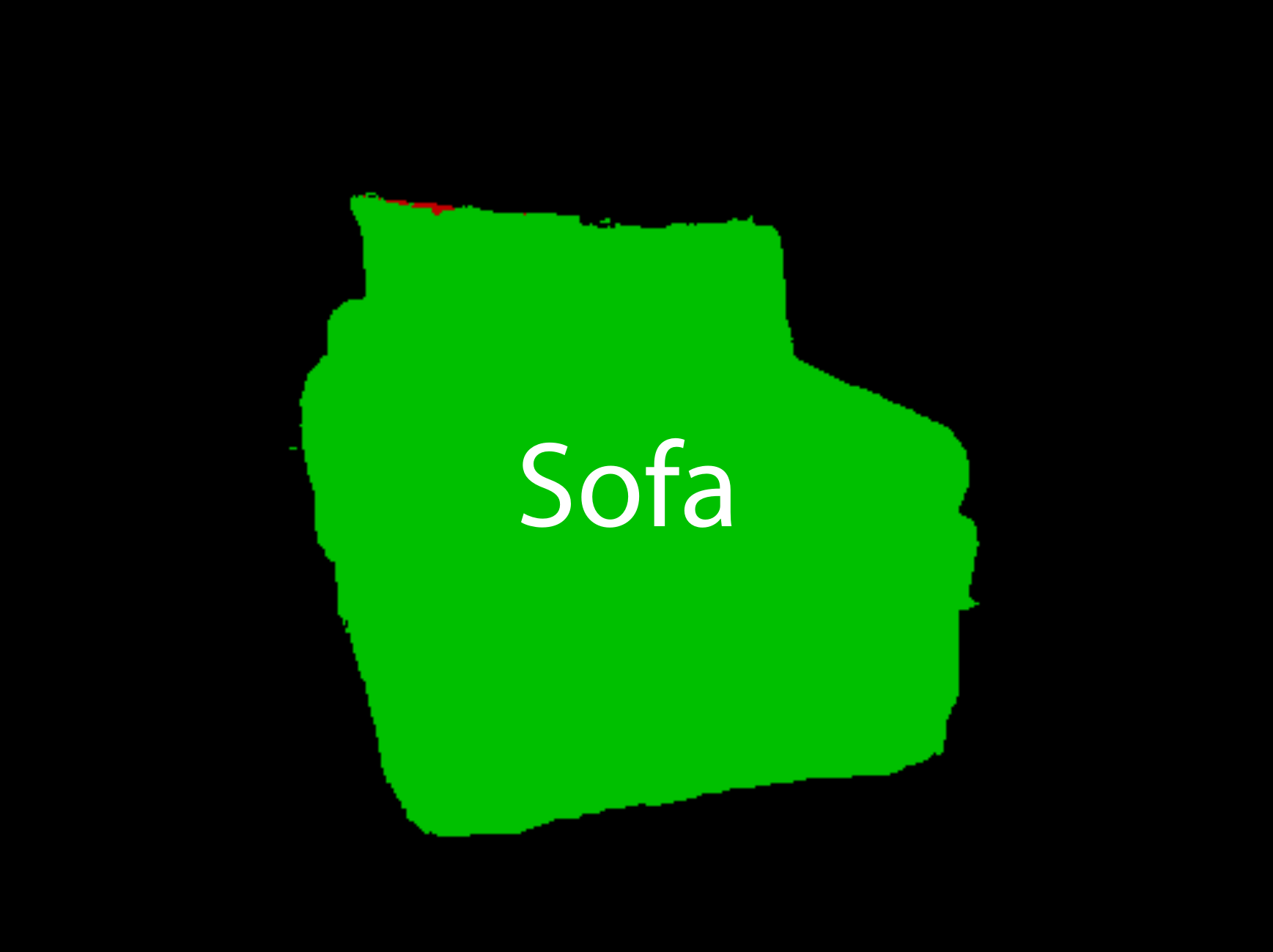}
  \end{subfigure}
  \begin{subfigure}[b]{0.15\linewidth}
    \includegraphics[width=\textwidth]{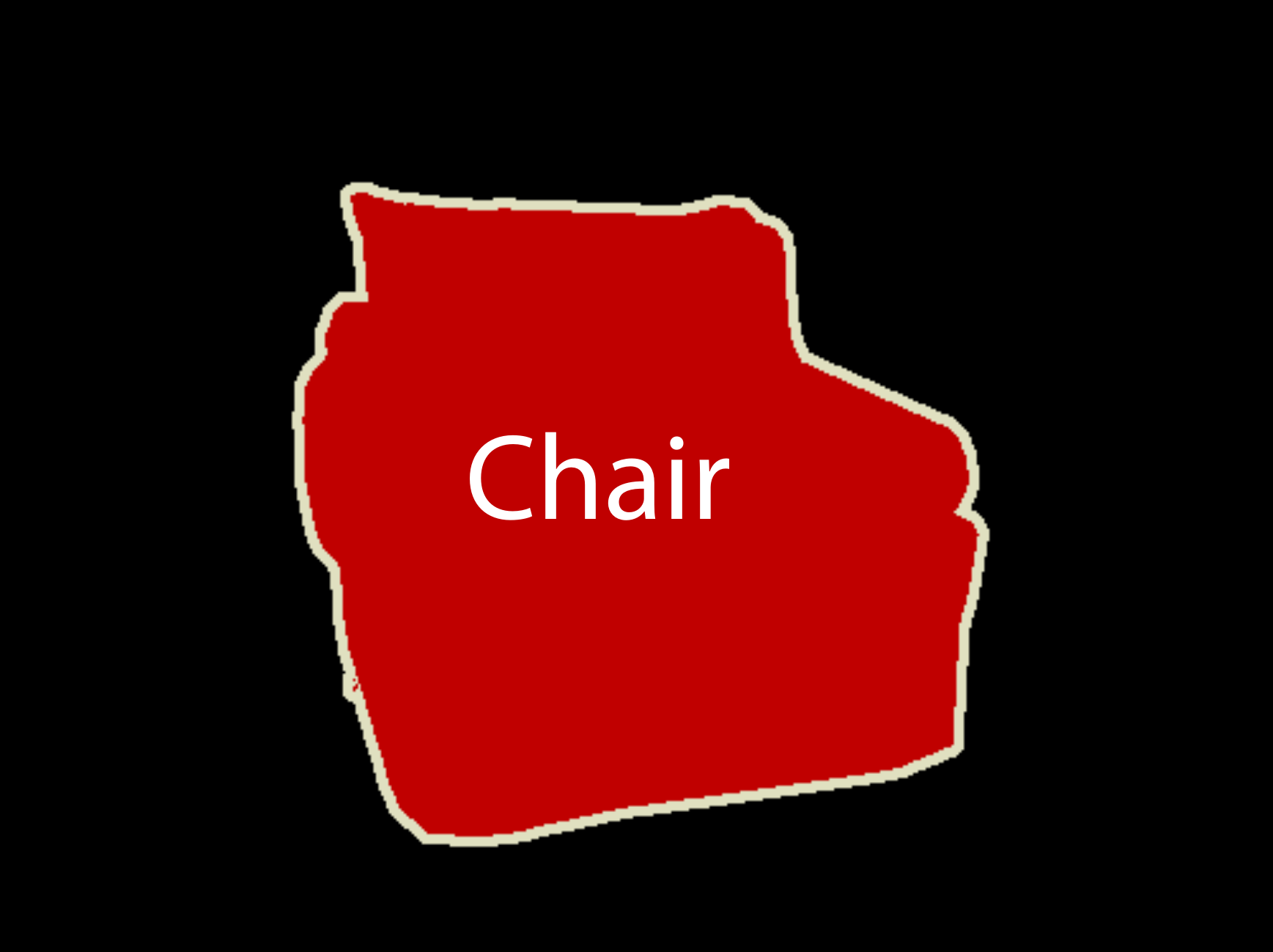}
  \end{subfigure}

  \begin{subfigure}[b]{0.15\linewidth}
    \includegraphics[width=\textwidth]{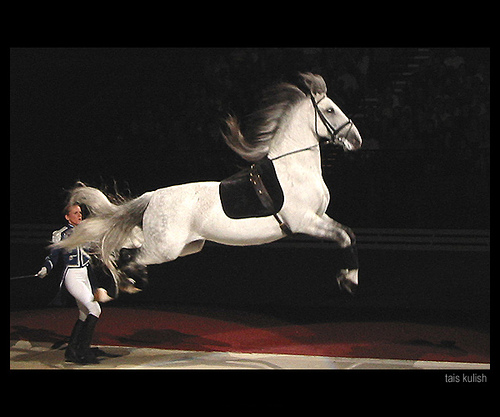}
    \captionsetup{labelformat=empty}
    \vspace{-0.2in}
    \caption{Image}
  \end{subfigure}
  \begin{subfigure}[b]{0.15\linewidth}
    \includegraphics[width=\textwidth]{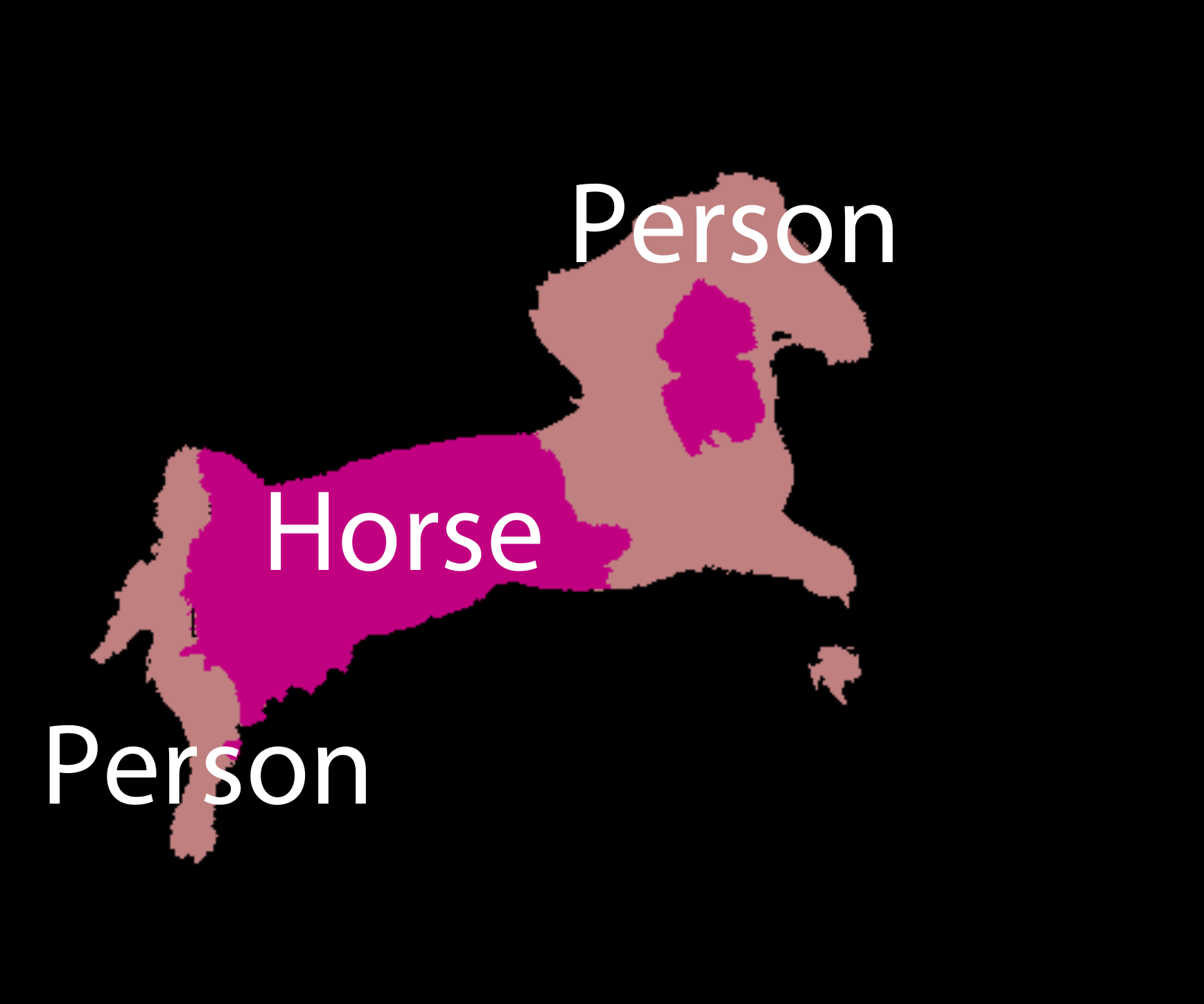}
    \captionsetup{labelformat=empty}
    \vspace{-0.2in}
    \caption{Our result}
  \end{subfigure}
  \begin{subfigure}[b]{0.15\linewidth}
    \includegraphics[width=\textwidth]{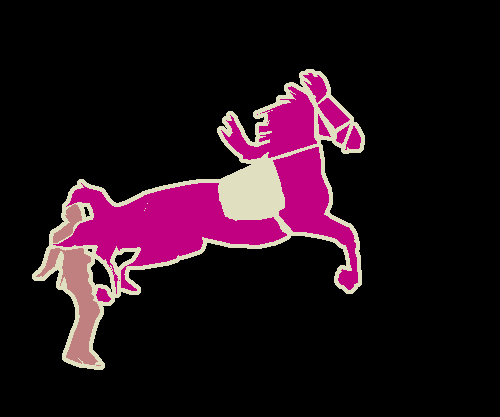}
    \captionsetup{labelformat=empty}
    \vspace{-0.2in}
    \caption{Ground truth}
  \end{subfigure}
  ~
  \begin{subfigure}[b]{0.15\linewidth}
    \includegraphics[width=\textwidth]{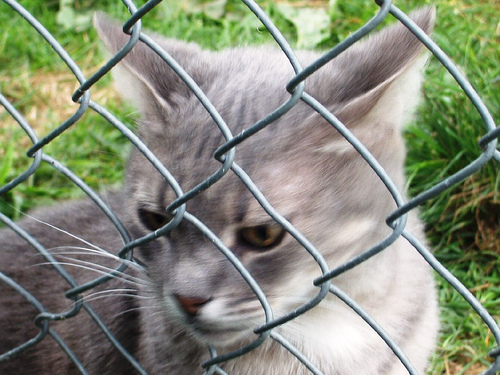}
    \captionsetup{labelformat=empty}
    \vspace{-0.2in}
    \caption{Image}
  \end{subfigure}
  \begin{subfigure}[b]{0.15\linewidth}
    \includegraphics[width=\textwidth]{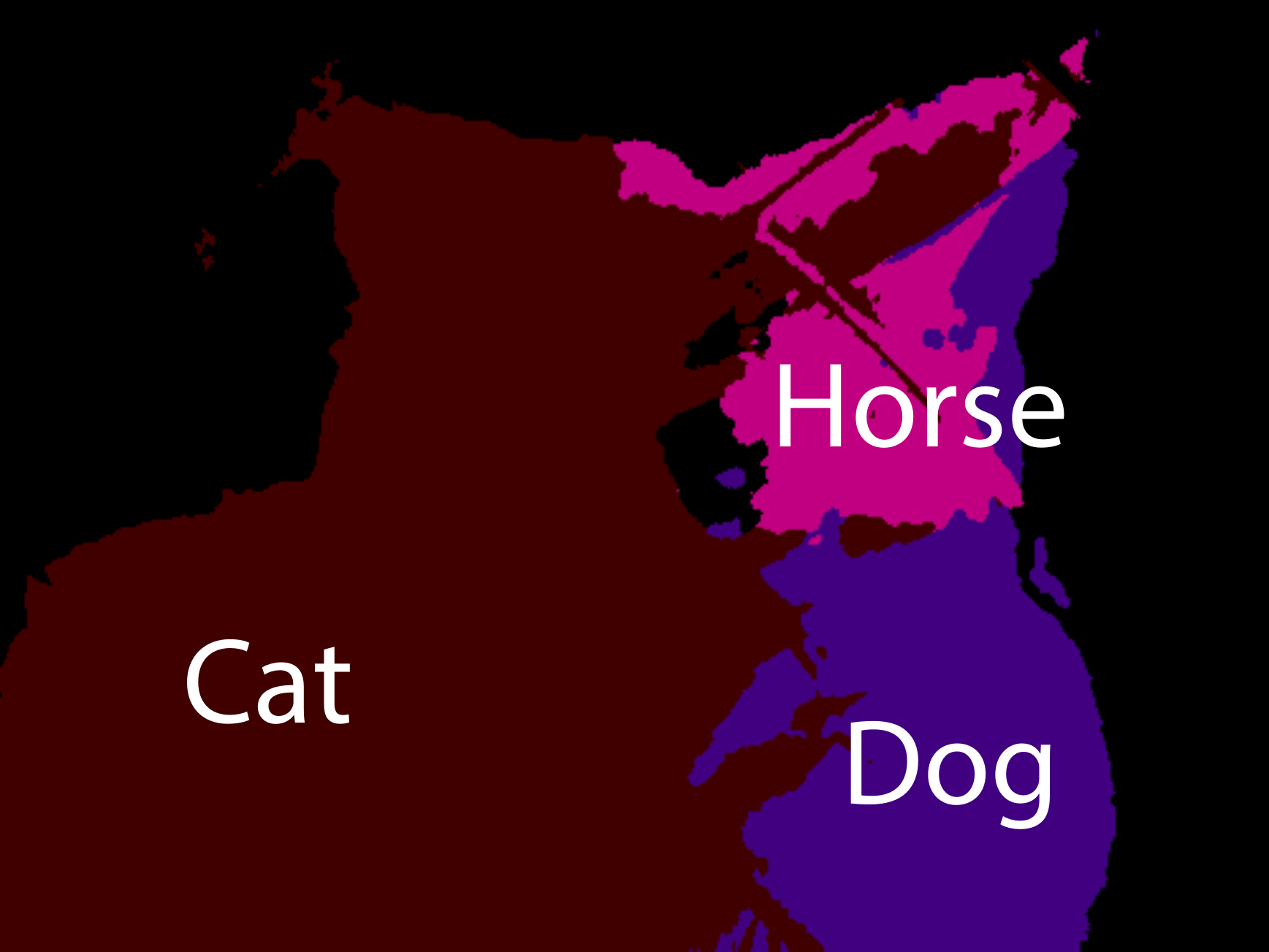}
    \captionsetup{labelformat=empty}
    \vspace{-0.2in}
    \caption{Our result}
  \end{subfigure}
  \begin{subfigure}[b]{0.15\linewidth}
    \includegraphics[width=\textwidth]{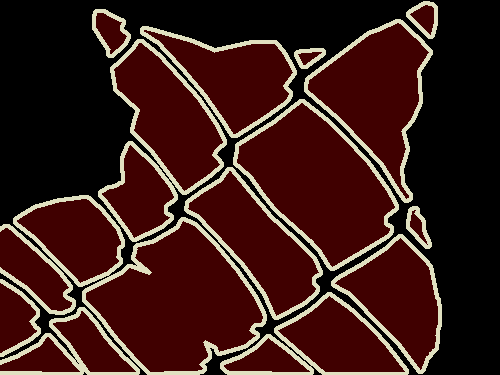}
    \captionsetup{labelformat=empty}
    \vspace{-0.2in}
    \caption{Ground truth}
  \end{subfigure}

  \vspace{-1mm}
  \caption{Failure cases from the VOC-2012 validation set. The most accurate architecture we trained (Context + CRF-RNN) performs poorly on these images.}
  \label{fig:failure}
  \end{figure}

State-of-the-art systems for semantic segmentation leave significant room for future advances.
Failure cases of our most accurate configuration are shown in Figure \ref{fig:failure}. We will release our code and trained models to support progress in this area.

\section*{Acknowledgements}

We thank Vibhav Vineet for proofreading, help with experiments, and related discussions. We are also grateful to Jonathan Long and the Caffe team for their feedback and for rapidly pulling our implementation into the Caffe library.

{\small
\bibliographystyle{iclr2016_conference}
   \setlength{\bibsep}{6pt}
   \linespread{1}\selectfont
\bibliography{dilation}
}

\newpage

\input{appendix}

\end{document}

%% file: appendix.tex
\begin{appendices}

\section{Urban Scene Understanding}

In this appendix, we report experiments on three datasets for urban scene understanding: the CamVid dataset~\citep{Brostow:2009:PRL}, the KITTI dataset \citep{GeigerLSU13}, and the new Cityscapes dataset~\citep{Cordts:2016:CVPR}. As the accuracy measure we use the mean IoU~\citep{Everingham2010}. We only train our model on the training set, even when a validation set is available. The results reported in this section do not use conditional random fields or other forms of structured prediction. They were obtained with convolutional networks that combine a front-end module and a context module, akin to the ``Front + Basic" network evaluated in Table \ref{tab:controlled}. The trained models can be found at \url{https://github.com/fyu/dilation}.

We now summarize the training procedure used for training the front-end module. This procedure applies to all datasets. Training is performed with stochastic gradient descent. Each mini-batch contains 8 crops from randomly sampled images. Each crop is of size $628\!\times\!628$ and is randomly sampled from a padded image. Images are padded using reflection padding. No padding is used in the intermediate layers. The learning rate is $10^{-4}$ and momentum is set to $0.99$. The number of iterations depends on the number of images in the dataset and is reported for each dataset below.

The context modules used for these datasets are all derived from the ``Basic" network, using the terminology of Table \ref{tab:layers}. The number of channels in each layer is the number of predicted classes $C$. (For example, $C\!=\!19$ for the Cityscapes dataset.) Each layer in the context module is padded such that the input and response maps have the same size. The number of layers in the context module depends on the resolution of the images in the dataset.
Joint training of the complete model, composed of the front-end and the context module, is summarized below for each dataset.

\subsection{CamVid}

We use the split of~\cite{SturgessALT09}, which partitions the dataset into 367 training images, 100 validation images, and 233 test images. 11 semantic classes are used. The images are downsampled to $640 \timess 480$.

The context module has 8 layers, akin to the model used for the Pascal VOC dataset in the main body of the paper. The overall training procedure is as follows. First, the front-end module is trained for 20K iterations. Then the complete model (front-end + context) is jointly trained by sampling crops of size $852 \timess 852$ with batch size 1. The learning rate for joint training is set to
$10^{-5}$ and the momentum is set to $0.9$.

Results on the CamVid test set are reported in Table~\ref{tab:camvid}. We refer to our complete convolutional network (front-end + context) as Dilation8, since the context module has 8 layers. Our model outperforms the prior work. This model was used as the unary classifier in the recent work of~\cite{Kundu2016}.

\begin{table}[htbp]
  \small
  \setlength{\tabcolsep}{5.3pt}
  \small
  \begin{tabular}{l||c|c|c|c|c|c|c|c|c|c|c||c}
     & \ver{Building} & \ver{Tree} & \ver{Sky} & \ver{Car} & \ver{Sign} & \ver{Road} & \ver{~~Pedestrian~~} &
    \ver{Fence} & \ver{Pole} & \ver{Sidewalk} & \ver{~Bicyclist~} & \ver{mean IoU} \\ \hline
ALE & 73.4 & 70.2 & \textbf{91.1} & 64.2 & 24.4 & 91.1 & 29.1 & 31.0 &
13.6 & 72.4 & 28.6 & 53.6 \\
SuperParsing & 70.4 & 54.8 & 83.5 & 43.3 & 25.4 & 83.4 & 11.6 & 18.3
& 5.2 & 57.4 & 8.9 & 42.0 \\
Liu and He & 66.8 & 66.6 & 90.1 & 62.9 & 21.4 & 85.8 & 28.0 & 17.8 &
8.3 & 63.5 & 8.5 & 47.2 \\
SegNet & 68.7 & 52.0 & 87.0 & 58.5 & 13.4 & 86.2 & 25.3 & 17.9 & 16.0
& 60.5 & 24.8 & 46.4 \\
DeepLab-LFOV & 81.5 & 74.6 & 89.0 & 82.2 & 42.3 & \textbf{92.2} & 48.4 & 27.2 & 14.3 & \textbf{75.4} & 50.1 & 61.6 \\
Dilation8 & \textbf{82.6} & \textbf{76.2} & 89.9 & \textbf{84.0} &
\textbf{46.9} & \textbf{92.2} & \textbf{56.3} & \textbf{35.8} &
\textbf{23.4} & 75.3 & \textbf{55.5} & \textbf{65.3}
\\ \hline
  \end{tabular}
  \caption{Semantic segmentation results on the CamVid dataset. Our model (Dilation8) is compared to ALE \citep{LadickyRKT09}, SuperParsing \citep{TigheLazebnik2013}, Liu and He \citep{LiuHe2015}, SegNet \citep{Badrinarayanan2015}, and the DeepLab-LargeFOV model~\citep{Chen2015ICLR}. Our model outperforms the prior work.}
  \label{tab:camvid}
\end{table}

\subsection{KITTI}

We use the training and validation split of~\cite{Ros:2015:WACV}: 100 training images and 46 test images. The images were all collected from the KITTI visual odometry/SLAM dataset. The image resolution is $1226 \timess 370$. Since the vertical resolution is small compared to the other datasets, we remove Layer 6 in Table~\ref{tab:layers}. The resulting context module has 7 layers. The complete network (front-end + context) is referred to as Dilation7.

The front-end is trained for 10K iterations. Next, the front-end and the context module are trained jointly. For joint training, the crop size is $900\timess 900$ and momentum is set to 0.99, while the other parameters are the same as the ones used for the CamVid dataset. Joint training is performed for 20K iterations.

The results are shown in Table~\ref{tab:kitti}. As the table demonstrates, our model outperforms the prior work.

\begin{table}[htbp]
\small
\setlength{\tabcolsep}{5pt}
\centering
\begin{tabular}{l||c|c|c|c|c|c|c|c|c|c|c||c}
 & \ver{Building} & \ver{Tree} & \ver{Sky} & \ver{Car} & \ver{Sign} & \ver{Road} & \ver{~~Pedestrian~~} &
\ver{Fence} & \ver{Pole} & \ver{Sidewalk} & \ver{Bicyclist} &
\ver{mean IoU} \\ \hline
Ros et al. & 71.8 & 69.5 & \textbf{84.4} & 51.2 & 4.2 & 72.4 & 1.7 &
32.4 & 2.6 & 45.3 & 3.2 & 39.9 \\
DeepLab-LFOV  & 82.8 & 78.6 & 82.4 & 78.0 & 28.8 & 91.3 & 0.0 &
39.4 & 29.9 & 72.4 & 12.9 & 54.2 \\
Dilation7 & \textbf{84.6} & \textbf{81.1} & 83 & \textbf{81.4} &
\textbf{41.8} & \textbf{92.9} & \textbf{4.6} & \textbf{47.1} & \textbf{35.2} &
\textbf{73.1} & \textbf{26.4} & \textbf{59.2} \\ \hline
\end{tabular}
\caption{Semantic segmentation results on the KITTI dataset. We compare our results to~\cite{Ros:2015:WACV} and to the DeepLab-LargeFOV model~\citep{Chen2015ICLR}. Our network (Dilation7) yields higher accuracy than the prior work.}
\label{tab:kitti}
\end{table}

\subsection{Cityscapes}


The Cityscapes dataset contains 2975 training images, 500 validation images, and 1525 test images~\citep{Cordts:2016:CVPR}. Due to the high image resolution ($2048 \timess 1024$), we add two layers to the context network after Layer 6 in Table~\ref{tab:layers}. These two layers have dilation 32 and 64, respectively. The total number of layers in the context module is 10 and we refer to the complete model (front-end + context) as Dilation10.

The Dilation10 network was trained in three stages. First, the front-end prediction module was trained for 40K iterations. Second, the context module was trained for 24K iterations on whole (uncropped) images, with learning rate $10^{-4}$, momentum $0.99$, and batch size 100. Third, the complete model (front-end + context) was jointly trained for 60K iterations on halves of images (input size
$1396 \timess 1396$, including padding), with learning rate $10^{-5}$, momentum $0.99$, and batch size
1.

\begin{figure}[t]
  \centering
  \includegraphics[width=\textwidth]{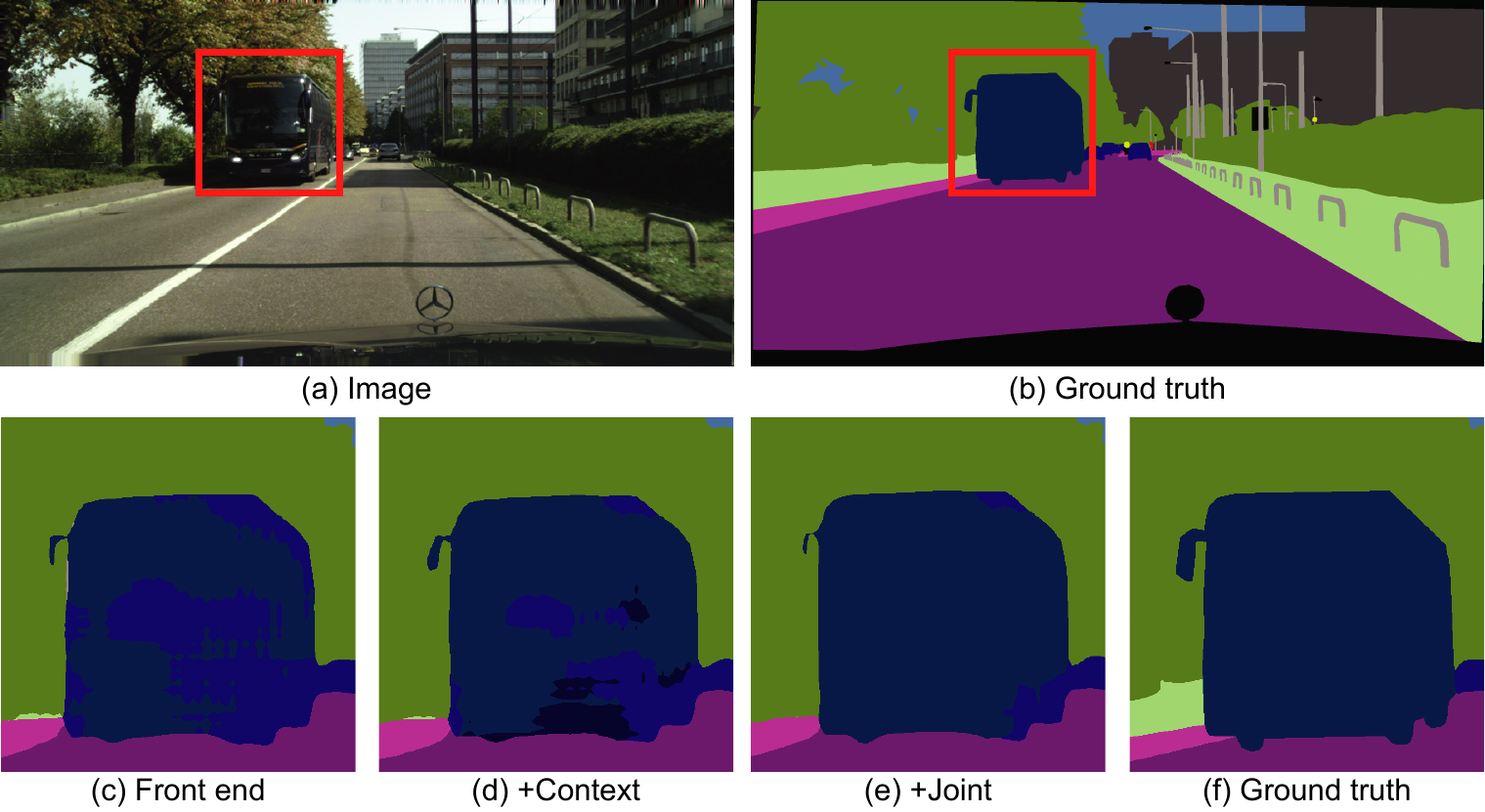}
  \caption{Results produced by the Dilation10 model after different training stages. (a) Input image. (b) Ground truth segmentation. (c) Segmentation produced by the model after the first stage of training (front-end only). (d) Segmentation produced after the second stage, which trains the context module. (e) Segmentation produced after the third stage, in which both modules are trained jointly.}
  \label{fig:cityscapes_stages}
\end{figure}

Figure~\ref{fig:cityscapes_stages} visualizes the effect of the training stages on the performance of the model. Quantitative results are given in Tables~\ref{tab:cityscape-class}
and~\ref{tab:cityscape-category}.

The performance of Dilation10 was compared to prior work on the Cityscapes dataset by~\cite{Cordts:2016:CVPR}. In their evaluation, Dilation10 outperformed all prior models~\citep{Cordts:2016:CVPR}. Dilation10 was also used as the unary classifier in the recent work of~\cite{Kundu2016}, which used structured prediction to increase accuracy further.

\begin{table}[htbp]
  \setlength{\tabcolsep}{1.5pt}
  \small
  \centering
  \renewcommand{\arraystretch}{1.1}
  \begin{tabular}{c|c|c|c|c|c|c|c|c|c|c|c|c|c|c|c|c|c|c||c}
    \ver{Road} & \ver{Sidewalk} & \ver{Building} & \ver{Wall} &
    \ver{Fence} & \ver{Pole} & \ver{Light} & \ver{Sign} & \ver{Vegetation} & \ver{Terrain} & \ver{Sky} & \ver{Person} & \ver{Rider}
    & \ver{Car} & \ver{Truck} & \ver{Bus} & \ver{Train} &
    \ver{~~Motorcycle~~} & \ver{Bicycle} & \ver{mean IoU}
    \\ \hline
    \multicolumn{20}{c}{Validation set} \\ \hline
    97.2 & 79.5 & 90.4 & 44.9 & 52.4 & 55.1 & 56.7 & 69 & 91 &
    58.7 & 92.6 & 75.7 & 50 & 92.2 & 56.2 & 72.6 & 54.3 & 46.2 & 70.1
    & 68.7 \\
    \hline
    \multicolumn{20}{c}{Test set} \\ \hline 
    97.6 & 79.2 & 89.9 & 37.3 & 47.6 & 53.2 & 58.6 & 65.2 & 91.8
    & 69.4 & 93.7 & 78.9 & 55 & 93.3 & 45.5 & 53.4 & 47.7 & 52.2 & 66
    & 67.1 \\
    \hline
  \end{tabular}
  \caption{Per-class and mean class-level IoU achieved by our model (Dilation10) on the Cityscapes dataset.}
  \label{tab:cityscape-class}
\end{table}

\begin{table}[htbp]
  \small
  \centering
  \renewcommand{\arraystretch}{1.1}
  \newcolumntype{C}[1]{>{\centering\let\newline\\\arraybackslash\hspace{0pt}}m{#1}}
  \setlength{\tabcolsep}{1.5pt}
  \begin{tabular}{C{1.6cm}|C{1.6cm}|C{1.6cm}|C{1.6cm}|C{1.8cm}|C{1.6cm}|C{1.6cm}||c}
    Flat & Nature & Object & Sky & Construction & Human & Vehicle &
    ~mean IoU~ \\ \hline
    \multicolumn{8}{c}{Validation set} \\ \hline
    98.2 & 91.4 & 62.3 & 92.6 & 90.7 & 77.6 & 91 & 86.3
    \\
    \hline
    \multicolumn{8}{c}{Test set} \\ \hline
    98.3 & 91.4 & 60.5 & 93.7 & 90.2 & 79.8 & 91.8 & 86.5
    \\
    \hline
  \end{tabular}
  \caption{Per-category and mean category-level IoU on the Cityscapes dataset.}
  \label{tab:cityscape-category}
\end{table}

\end{appendices}